*

# Human-Machine Shared Control Approach for the Takeover of CACC

Haoran Wang, Zhexi Lian, Zhenning Li, Jiawei Wang, Arno Eichberger, *Member, IEEE*,

Jia Hu, *Senior Member, IEEE*, Yongyu Chen, and Yongji Gao

***Abstract*—Cooperative Adaptive Cruise Control (CACC) often requires human takeover for tasks such as exiting a freeway. Direct human takeover can pose significant risks, especially given the close-following strategy employed by CACC, which might cause drivers to feel unsafe and execute hard braking, potentially leading to collisions. This research aims to develop a CACC takeover controller that ensures a smooth transition from automated to human control. The proposed CACC takeover maneuver employs an indirect human-machine shared control approach, modeled as a Stackelberg competition where the machine acts as the leader and the human as the follower. The machine guides the human to respond in a manner that aligns with the machine's expectations, aiding in maintaining following stability. Additionally, the human reaction function is integrated into the machine's predictive control system, moving beyond a simple "prediction-planning" pipeline to enhance planning optimality. The controller has been verified to i) enable a smooth takeover maneuver of CACC; ii) ensure string stability in the condition that the platoon has less than 6 CAVs and human control authority is less than 40%; iii) enhance both perceived and actual safety through machine interventions; and iv) reduce the impact on upstream traffic by up to 60%.**

***Index Terms*—CACC, Platoon takeover, human-machine shared control, game-based model predictive control**

## I. Introduction

Cooperative Adaptive Cruise Control (CACC) is deemed to be a promising technology in the field of cooperative automation [1]. By enabling conventional Adaptive Cruise Control (ACC) with communication and cooperation, CACC is capable of maneuvering Cooperative and Autonomous Vehicles (CAVs) tightly following the preceding vehicle. CACC has been verified to double road capacity [2, 3], save 10%-20% of energy [4], and reduce 14% of carbon dioxide emission [5].

Although bearing these merits, CACC is not yet widely promoted in field implementations. One of the main concerns is the safety risks during the takeover maneuver. In the field implementation of CACC, a human driver would take over the control authority from time to time, including conducting a lane change and exiting a highway. Control authority transitions from automation to human drivers are broadly classified into two modes: automation-initiated (e.g., sensor limitations) and driver-initiated (e.g., lane changes) [6]. Automation-initiated transitions carry heightened risks, as drivers often face physiological and cognitive delays when abruptly regaining situational awareness [7]. For example, during transitions from ACC to manual control, studies report sub-5-second delays for drivers to regain full control [8, 9]. This delay encompasses reaction time, attention recovery, and cognitive recalibration [10, 11]. While driver-initiated transitions are inherently lower-risk (due to retained situational awareness), they are not without hazards. Abrupt steering corrections during voluntary overrides can conflict with the system's trajectory, increasing lateral deviation risks [12]. Thus, adaptive assistance systems—such as predictive haptic guidance or shared control frameworks—are critical for harmonizing human-machine coordination during both transition types.

Furthermore, these challenges are further amplified in CACC's cooperative driving context. According to field tests, the following headway in a CACC platoon could be reduced to 0.5 seconds, much shorter than the 1-2 seconds headway of the conventional ACC and Human-driven Vehicle (HV) [13]. The close-following strategy of CACC is usually perceived as unsafe for a human driver. Suffering this unsafe feeling, the human driver tends to conduct hard braking without the

*This paper is partially supported by National Key R&D Program of China (2022YFF0604905), National Natural Science Foundation of China (Grant No. 52302412 and 52372317), Yangtze River Delta Science and Technology Innovation Joint Force (No. 2023CSJGG0800), the Fundamental Research Funds for the Central Universities, Tongji Zhongte Chair Professor Foundation (No. 000000375-2018082), Shanghai Sailing Program (No. 23YF1449600), and the Science Fund of State Key Laboratory of Advanced Design and Manufacturing Technology for Vehicle (No. 32215011). *(Corresponding author: Jia Hu.)*

H. Wang, Z. Lian, and J. Hu are with Key Laboratory of Road and Traffic Engineering of the Ministry of Education, Tongji University, Shanghai 201804, China. (e-mail: wang_haoran@tongji.edu.cn; zhexi_lian@tongji.edu.cn; hujia@tongji.edu.cn).

Z. Li is with State Key Laboratory of Internet of Things for Smart City, Departments of Civil and Environmental Engineering and Computer and Information Science, University of Macau, Avenida da Universidade Taipa, Macau, China. (e-mail: zhenningli@um.edu.mo).

J. Wang is with the Department of Civil and Environmental Engineering, University of Michigan, Ann Arbor, USA. (email: jiawe@umich.edu).

A. Eichberger is with Institute of Automotive Engineering, Graz University of Technology, Graz, A-8010, Austrian. (e-mail: arno.eichberger@tugraz.at).

Yongyu Chen is with Li Auto, Yutian Building, No. 111 South Yutian Road, Jiading District, Shanghai, P.R.China. (email: yongyu.chen@tum.de)

Y. Gao is with Chery Automobile Co., Ltd. Institute of Automotive Engineering Research and Development Kaiyang Laboratory Yao Guang Laboratory Cluster Information Perception Technology Laboratory, No. 21 Fengming Hubei Road, Longshan Street, Economic and Technological Development Zone, Wuhu City, Anhui Province, 241008, China (e-mail: gaoyongji@mychery.com)

consideration of his back. The human driver's stress behavior would cause collisions. Therefore, it is quite crucial to ensure the safety of CACC's takeover.

However, as far as we know, there are still no studies developing takeover controllers for CACC. Past studies mostly focus on the regular operation of CACC [14-17]. However, in takeover maneuvers, these CACC controllers are not compatible anymore. A coexistent strategy for both human and machine shall be developed. A smooth control authority transition from machine to human shall be realized. Nevertheless, previous studies have proposed various methods for the takeover of other systems, using haptic shared control [18] and rule-based shared control [19]. These studies would provide great inspiration for the takeover of CACC systems.

In recent years, human-machine shared control has been a hot topic in the field of automated driving [20]. Currently, human-machine shared driving has been widely adopted and validated in ADAS systems on production vehicles, like Tesla Model X and Lincoln Z-type. The human-machine shared control method can be typically divided into two categories: direct shared control and indirect shared control [21]. The direct shared control imposes the machine's commands on the steering wheel via haptic feedback [22]. This type of method cannot be adapted to the CACC takeover scenario, since longitudinal shared control cannot be realized via throttle or brake pedals' direct feedback. To address this challenge, the indirect shared control method is proposed to generate machine inputs and fuse with human commands before execution [23, 24]. This method enables a gradual authority transition from machine to human. It has been adopted in a lot of previous studies on takeover control of automation [21, 25-27]. These studies verified the effectiveness of gradual authority transition on driving simulators. Hence, drivers' acceptance may not be a great issue in applications.

In this paper, in response to the practical requirement of implementing CACC, this study introduces a novel CACC takeover controller based on the human-machine indirect shared control approach. The proposed controller bears the following contributions:

• **Enabling a smooth takeover maneuver of CACC**: The controller employs a human-machine indirect shared control approach to facilitate a seamless transition from automated to human control. *Machine interventions help mitigate human drivers' aggressive responses*, such as hard braking, which are often triggered by the close-following strategy of CACC. By gradually transferring control authority from the machine to the human, the controller ensures a smooth takeover process.

• **Achieving planning optimality in human-machine shared control**: Unlike the traditional "prediction-planning" pipeline, *this research integrates the human reaction function into the machine's predictive control system*, ensuring planning optimality. The human-machine shared control is modeled as a Stackelberg competition, with the machine's planning framed as a Game-based Model Predictive Control (GMPC) problem. By incorporating the human reaction function into the system dynamics, the machine's commands are derived as optimal solutions considering human reaction rationales.

• **Ensuring string stability within a specific Operational Design Domain (ODD)**: The proactive interventions of the machine enable the ego-vehicle to anticipate the actions of the platoon leader by utilizing the future actions of preceding vehicles. Allocating partial control authority to the machine stabilizes the CACC system, preventing oscillations from propagating along the string. *The ODD for maintaining string stability has been quantified* based on various factors of authority allocation between the human and the machine.

• **Enhancing driving safety and reduced traffic impacts**: The proposed controller's proactive commands reduce following gap oscillations and acceleration magnitude. Hence, it enhances both perceived safety and actual safety. Additionally, the maintenance of string stability minimizes oscillations experienced by the last CAV in the platoon, thereby mitigating adverse effects on upstream traffic.

## II. METHODOLOGY

The proposed CACC takeover controller focuses on the automation-initiated authority transition from machine to human driver. It follows the human-machine shared control framework. In the system, each CAV would make a Stackelberg competition with its human driver. Human reaction function is modeled and considered in the machine's planning. The machine's planned commands are fused with the human driver's commands for execution.

### *A. System structure*

The system structure of the proposed controller is shown in Fig. 1. The CACC platoon follows a Predecessor-Following (PF) topology. CAVs within the platoon are controlled in a distributed approach. Each CAV is under the shared control of human and machine. From the human side, the human driver wants to switch his role from supervising to driving. The human monitors the preceding vehicle and makes throttle or brake commands to approach the desired path. From the machine side, the machine controller obtains status information on the environment via onboard perception and communication. The machine outputs commands to maintain the following stability. Obtaining commands from human and machine, a shared control law is designed to generate final adjusted commands to steer-by-wire chassis.

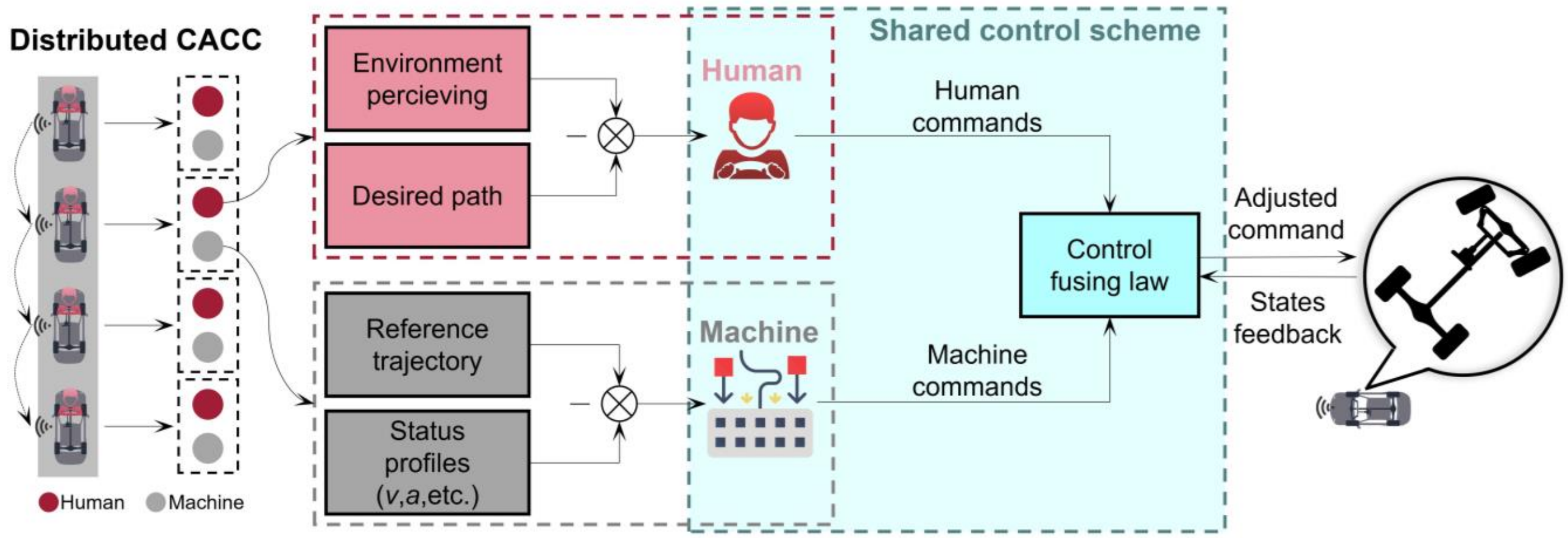

Fig. 1. System framework for CACC takeover.

A human and machine indirect shared control scheme [23] is adopted for the takeover maneuver of CACC, as illustrated in Fig. 2. The system is applicated within a rolling horizon framework [28, 29]. The scheme mainly consists of three modules: human driver, human modeling, machine controller, and control fusing.

• **Human driver:** The human driver would manipulate the throttle and brake to control the vehicle. Human commands are transformed into commanded acceleration as the input of control fusing.

• **Human modeling**: A Model Predictive Control (MPC) based modeling method is proposed to model the human's reaction to the environment. The modeling outputs a human reaction function, which reveals the law of human driver's behavior. The human reaction function would be considered in the formulation of human-machine system dynamics in machine planning.

• **Machine controller:** The machine controller is formulated into a GMPC problem. The machine would conduct a Stackelberg competition with the human. In this competition, machine is the leader, and the human is the follower. The machine is with the objective of leading the human to react as the machine expects. Hence, a human and machine shared system dynamics model is formulated by introducing the human reaction function into the dynamics model. In this way, the machine is capable of leading the human as expected to realize a stable following.

• **Control fusing:** This module is proposed to generate the final commands. A typical control fusing function is adopted. It considers the ODD of ensuring string stability and human driver's status to allocate control authorities between human and machine. Specifically, the system observes human driver's control commands. When drivers initiate transition, their immediate overrides are granted. In automation-initiated transitions, system fuses with machine commands when smooth human commands are observed. 100% of control authority when human drivers' control commands are not smooth anymore. Based on these information, weighting factors are decided to fuse human and machine commands. Final adjusted commands are output for local execution.

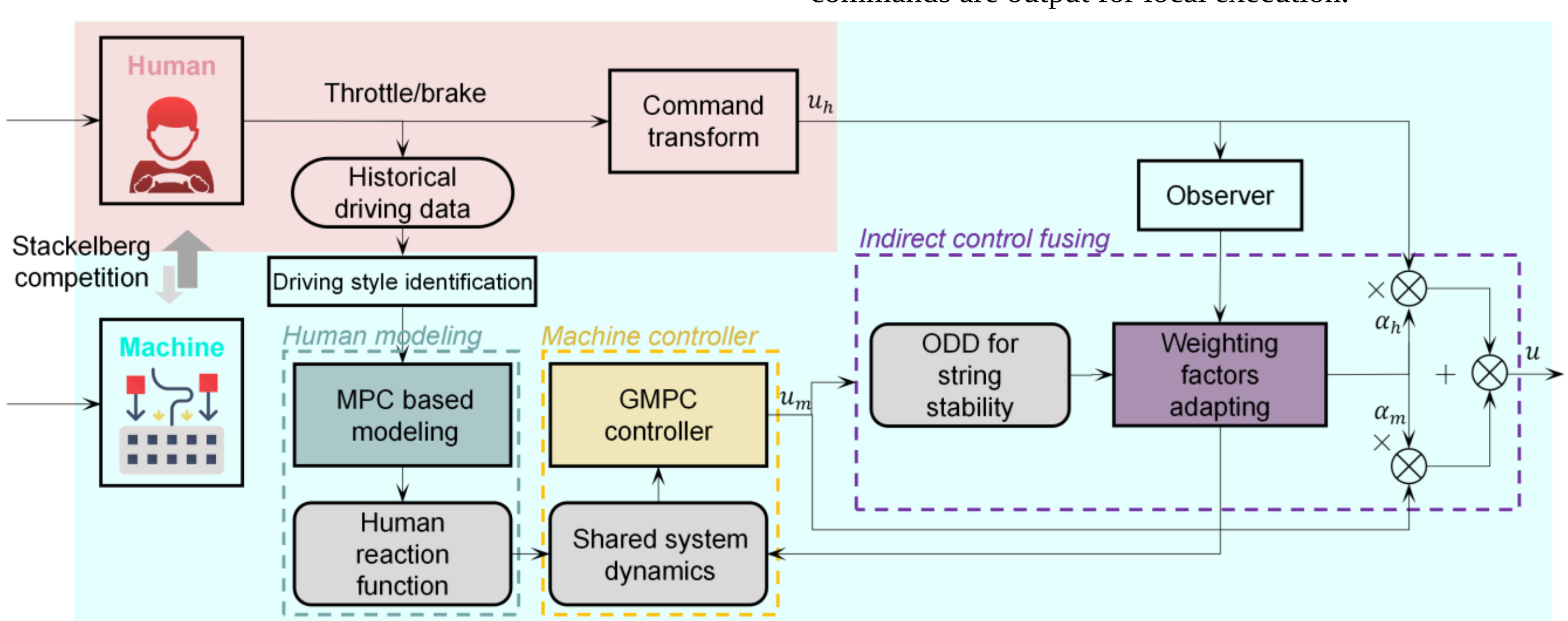

Fig. 2. Stackelberg competition based indirect shared control scheme.

*B. Definition and Declaration*

Following the distributed CACC system structure, each CAV within the platoon is controlled by a human and machine shared controller. Hence, the following formulation only presents a distributed vehicle controller. State and control vectors of a human-machine shared control system are defined in the following.

State vector $\boldsymbol{x}_k$ at control step $k$ is defined as follows.

$$\boldsymbol{x}_k \stackrel{\text{def}}{=} [\Delta v_k, g_k]^T \tag{1}$$

where $\Delta v_k$ is the speed difference between two vehicles in a CACC platoon; $g_k$ is the following gap between two vehicles in a CACC platoon. $\Delta v_k$ and $g_k$ are defined as follows.

$$\Delta v_k \stackrel{\text{def}}{=} v_k - v_k^p \tag{2}$$

$$g_k \stackrel{\text{def}}{=} x_k^p - x_k \tag{3}$$

where $v_k$ is the vehicle speed in a platoon; $v_k^p$ is the speed of the preceding vehicle; $x_k^p$ is the longitudinal position of the preceding vehicle; $x_k$ is the longitudinal vehicle position in a platoon.

The desired state is defined as Eqs. (4) and (5). $\boldsymbol{x}_{h,k}^{ref}$ is the desired state of human driver. $\boldsymbol{x}_{m,k}^{ref}$ is the desired state of machine controller. When human driver takes over the authority, it generally expects a greater following gap $g_{h,k}^{ref}$, and ignores the following stability. The machine controller expects to maintain the stable following status.

$$\boldsymbol{x}_{h,k}^{ref} \stackrel{\text{def}}{=} \left[\sim, g_{h,k}^{ref}\right]^T \tag{4}$$

$$\boldsymbol{x}_{m,k}^{ref} \stackrel{\text{def}}{=} [0, \sim]^T \tag{5}$$

According to the indirect shared control scheme, the control vector $\boldsymbol{u}_k$ is the fusion of human's control vector $\boldsymbol{u}_{h,k}$ and machine's control vector $\boldsymbol{u}_{m,k}$, as shown in Eq. (6).

$$\boldsymbol{u}_k \stackrel{\text{def}}{=} \alpha_{h,k}\boldsymbol{u}_{h,k} + \alpha_{m,k}\boldsymbol{u}_{m,k} \tag{6}$$

$$\boldsymbol{u}_{h,k} \stackrel{\text{def}}{=} a_{h,k} \tag{7}$$

$$\boldsymbol{u}_{m,k} \stackrel{\text{def}}{=} a_{m,k} \tag{8}$$

where $\alpha_{h,k}$ is the authority allocation factor of human's command; $\alpha_{m,k}$ is the authority allocation of machine's command. Typically, $\alpha_{h,k} + \alpha_{m,k} = 1$ to avoid outliers. $a_{h,k}$ is the human's commanded acceleration; $a_{m,k}$ is the machine's commanded acceleration. The authority transition duration could be manually adjusted to suit various real-world scenarios. For example, the driver-initiated transition could use a short transition duration, since the driver has already anticipated the transition. On the contrary, the automation-initiated transition could be slower to adapt to driver's readiness.

### C. Human-machine shared system dynamics model

Derivatives of state variables in Eq. (1) are as follows.

$$\frac{d\Delta v(t)}{dt} = a^p(t) - a(t) \tag{9}$$

$$\frac{dg(t)}{dt} = \Delta v(t) \tag{10}$$

where $a^p$ is the speed of the preceding vehicle.

Formulating the Eqs. (9) and (10) into a state-space form as follows [30].

$$\dot{\boldsymbol{x}}(t) = \boldsymbol{A}'\boldsymbol{x}(t) + \boldsymbol{B}'\boldsymbol{u}(t) + \boldsymbol{C}'a^p(t) \tag{11}$$

where $\boldsymbol{A}' \stackrel{\text{def}}{=} \begin{bmatrix} 0 & 0 \\ 1 & 0 \end{bmatrix}$, $\boldsymbol{B}' \stackrel{\text{def}}{=} \begin{bmatrix} -1 \\ 0 \end{bmatrix}$, $\boldsymbol{C}' \stackrel{\text{def}}{=} \begin{bmatrix} 1 \\ 0 \end{bmatrix}$.

The continuous formulation of system dynamics shall be discretized since vehicle control is generally implemented in a discrete domain. Applying the forward Euler method, the control input is assumed to be constant within each update interval $\Delta t$: $\boldsymbol{u}(t) = \boldsymbol{u}_k$ where $k$ is the control step index and $k\Delta t \le t < (k+1)\Delta t$. The discretization is close to the original continuous form since the control interval is generally sufficiently small [31]. The continuous system dynamics could then be discretized as follows.

$$\boldsymbol{x}_{k+1} = \boldsymbol{A}\boldsymbol{x}_k + \boldsymbol{B}_{h,k}\boldsymbol{u}_{h,k} + \boldsymbol{B}_{m,k}\boldsymbol{u}_{m,k} + \boldsymbol{C}a_k^p \quad , \quad k \in [0, K) \tag{12}$$

where $\boldsymbol{A} \stackrel{\text{def}}{=} \boldsymbol{I} + \boldsymbol{A}'\Delta t$, $\boldsymbol{B} \stackrel{\text{def}}{=} \boldsymbol{B}'\Delta t$, $\boldsymbol{B}_{h,k} \stackrel{\text{def}}{=} \boldsymbol{B}\alpha_{h,k}$, $\boldsymbol{B}_{m,k} \stackrel{\text{def}}{=} \boldsymbol{B}\alpha_{m,k}$, and $\boldsymbol{C} = \boldsymbol{C}'\Delta t$. $K$ is the control horizon.

It should be noted that the vehicle state updating rule is different from the view of human and machine. Each machine controller is capable of obtaining the future actions of the preceding CAV via communication within the platoon. The human driver can only perceive the current state of the preceding vehicle. Accurate future actions are not available for human. Hence, from the view of human, $\boldsymbol{C}a_k^p$ is not obtained in the dynamics model (12).

### D. Human driving modeling

The human driving modeling is based on human factors. Human driver is regarded as a boundedly rational but goal-directed decision maker according to the bounded optimality theory [32]. Human drive a vehicle to best achieve their personalized objectives, based on their experience and a real-time prediction of environment. This behaving logic accords with MPC theory [33-35]. The cost function in MPC could reflect human's driving objective. The system dynamics in MPC could be the modeling of human experience. The rolling horizon implementation framework of MPC agrees with human's prediction-action behaving logic. Therefore, MPC modeling method could incorporate human's behavior factors. Hence, an MPC framework is adopted to model human driving in this research. The reaction function serves as a predictive behavioral model that solves the MPC problem to anticipate human driver actions over a finite time horizon. Specifically, it captures driver's tactical control prediction, like a time-varying acceleration profile.

#### 1) MPC-based human driving model

The human driving model could be formulated into the form of the LQR problem as follows. It aims at finding the optimal control series $\boldsymbol{u}_{h,k}\big|_{k=1}^{k=K}$ to minimize $J_h$, which is a surrogate objective function of human's actual objective function [36]. Although $J_h$ cannot accurately reveal the rationale of how human driving, it could reflect the state changing of human driving from a global view. Human's inconsistent driving behavior and control lagging could be considered by modeling the state changing.

$$\min_{\boldsymbol{u}_{h,k}|_{k=1}^{k=K}} (J_h) = \sum_{k=1}^{k=K-1} \Phi_{h,k}\left(\boldsymbol{x}_k, \boldsymbol{u}_{h,k}, \boldsymbol{x}_{h,k}^{ref}\right) + \Psi_{h,K}\left(\boldsymbol{x}_k, \boldsymbol{x}_{h,k}^{ref}\right) \tag{13}$$

$$\text{s.t. vehicle dynamics: } f_h\left(\boldsymbol{x}_k, \boldsymbol{u}_{h,k}, \boldsymbol{u}_{m,k}\right) = 0 \tag{14}$$

where $\Phi_{h,k}$ is the running cost of human driving; $\Psi_{h,K}$ is the terminal cost of human driving; $f_h$ is the vehicle dynamics. The three functions are formulated as follows.

$$\Phi_{h,k} \overset{\text{def}}{=} \frac{1}{2}\left(\boldsymbol{x}_k - \boldsymbol{x}_{h,k}^{ref}\right)^T \boldsymbol{Q}_{h,k}\left(\boldsymbol{x}_k - \boldsymbol{x}_{h,k}^{ref}\right) + \frac{1}{2}\boldsymbol{u}_{h,k}{}^T R_h \boldsymbol{u}_{h,k} \tag{15}$$

$$\Psi_{h,K} \overset{\text{def}}{=} \frac{1}{2}\left(\boldsymbol{x}_K - \boldsymbol{x}_{h,K}^{ref}\right)^T \boldsymbol{Q}_{h,K}\left(\boldsymbol{x}_K - \boldsymbol{x}_{h,K}^{ref}\right) \tag{16}$$

$$f_h = -\boldsymbol{x}_{k+1} + \boldsymbol{A}\boldsymbol{x}_k + \boldsymbol{B}_{h,k}\boldsymbol{u}_{h,k} + \boldsymbol{B}_{m,k}\boldsymbol{u}_{m,k} \tag{17}$$

where $\boldsymbol{Q}_{h,k} = \text{diag}\left[q_{h,k}^v, q_{h,k}^g\right]$. $q_{h,k}^v$, $q_{h,k}^g$, and $\boldsymbol{R}_{h,k}$ are weighting factors of speed error cost, gap error cost, and control effort cost, respectively. They are all positive. The vehicle dynamics model is obtained from Eq. (12) after removing the item of the preceding vehicle's future actions $\boldsymbol{C}a_k^p$.

*2) Human reaction function*

The human reaction function is the optimal control law of the problem (13). The control law is deduced based on the dynamic programming approach, as shown in ***Theorem 1***.

***Theorem 1***: The human's action function is as follows.

$$\boldsymbol{u}_{h,k} = \boldsymbol{K}_{h,k}\boldsymbol{x}_k + \boldsymbol{P}_{h,k}\boldsymbol{u}_{m,k} + \boldsymbol{S}_{h,k} \tag{18}$$

where

$$\boldsymbol{H}_{h,k} \overset{\text{def}}{=} \left(\boldsymbol{R}_{h,k} + \boldsymbol{B}_{h,k}{}^T\boldsymbol{D}_{k+1}\boldsymbol{B}_{h,k}\right)^{-1} \tag{19}$$

$$\boldsymbol{K}_{h,k} \overset{\text{def}}{=} -\boldsymbol{H}_{h,k}\boldsymbol{B}_{h,k}{}^T\boldsymbol{D}_{k+1}\boldsymbol{A} \tag{20}$$

$$\boldsymbol{P}_{h,k} \overset{\text{def}}{=} -\boldsymbol{H}_{h,k}\boldsymbol{B}_{h,k}{}^T\boldsymbol{D}_{k+1}\boldsymbol{B}_{m,k} \tag{21}$$

$$\boldsymbol{S}_{h,k} \overset{\text{def}}{=} -\boldsymbol{H}_{h,k}\boldsymbol{B}_{h,k}{}^T\boldsymbol{F}_{k+1} \tag{22}$$

$$\boldsymbol{N}_k \overset{\text{def}}{=} \boldsymbol{A} + \boldsymbol{B}_{h,k}\boldsymbol{K}_{h,k} \tag{23}$$

$$\boldsymbol{M}_{h,k} \overset{\text{def}}{=} -\left(\boldsymbol{K}_{h,k}{}^T\boldsymbol{R}_{h,k} + \boldsymbol{N}_k{}^T\boldsymbol{D}_{k+1}\boldsymbol{B}_{h,k}\right)\boldsymbol{H}_{h,k}\boldsymbol{B}_{h,k}{}^T \tag{24}$$

where $\boldsymbol{D}_k$, and $\boldsymbol{F}_k$ are concomitant coefficient matrices that could be computed in reverse order:

$$\boldsymbol{D}_k = \boldsymbol{Q}_h + \boldsymbol{K}_{h,k}{}^T\boldsymbol{R}_{h,k}\boldsymbol{K}_{h,k} + \boldsymbol{N}_k{}^T\boldsymbol{D}_{k+1}\boldsymbol{N}_k \tag{25}$$

$$\boldsymbol{F}_k = -\boldsymbol{Q}_h\boldsymbol{x}_{h,k}^{ref} + \left(\boldsymbol{M}_{h,k} + \boldsymbol{N}_k{}^T\right)\boldsymbol{D}_{k+1}\boldsymbol{B}_{m,k}\boldsymbol{u}_{m,k} + \left(\boldsymbol{M}_{h,k} + \boldsymbol{N}_k{}^T\right)\boldsymbol{F}_{k+1} \tag{26}$$

where $\boldsymbol{D}_K = \boldsymbol{Q}_h$ and $\boldsymbol{F}_K = -\boldsymbol{Q}_h\boldsymbol{x}_{h,K}^{ref}$.

Proof:

Based on the dynamic programming logic, the cost-to-go function of problem (13) is defined as $V(\boldsymbol{x}_k)$ in Eq. (27) as follows.

$$V(\boldsymbol{x}_k) = \min_{\boldsymbol{u}_{h,i}|_{i=k}^{i=K}} \sum_{i=k}^{i=K-1} \Phi_{h,i} + V(\boldsymbol{x}_K) \tag{27}$$

$$V(\boldsymbol{x}_K) = \Psi_{h,K} \tag{28}$$

where $V(\boldsymbol{x}_k)$ is the cost-to-go function at the step $k$.

Based on the Bellman's optimality principle, the optimal solution could be obtained by iteratively optimizing $V(\boldsymbol{x}_k)$ from the step $K-1$ to the step 1. At the step $k$, the cost-to-go function $V(\boldsymbol{x}_k)$ is assumed be fit into a second-order form as follows.

$$V(\boldsymbol{x}_k) = \Phi_{h,k} + \frac{1}{2}\boldsymbol{x}_{k+1}{}^T\boldsymbol{D}_{k+1}\boldsymbol{x}_{k+1} + \boldsymbol{x}_{k+1}{}^T\boldsymbol{F}_{k+1} + \boldsymbol{G}_{k+1} \tag{29}$$

where $\boldsymbol{D}$, $\boldsymbol{F}$, and $\boldsymbol{G}$ are coefficient matrices that would be inferred in the following.

The derivative of $V(\boldsymbol{x}_k)$ is computed by Eq. (30).

$$\begin{aligned}\frac{\partial V(\boldsymbol{x}_k)}{\partial \boldsymbol{u}_{h,k}} &= \boldsymbol{R}_{h,k}\boldsymbol{u}_{h,k} + \frac{\partial \boldsymbol{x}_{k+1}{}^T}{\partial \boldsymbol{u}_{h,k}}\boldsymbol{D}_{k+1}\boldsymbol{x}_{k+1} + \frac{\partial \boldsymbol{x}_{k+1}{}^T}{\partial \boldsymbol{u}_{h,k}}\boldsymbol{F}_{k+1} \\ &= \boldsymbol{R}_{h,k}\boldsymbol{u}_{h,k} + \boldsymbol{B}_{h,k}{}^T\boldsymbol{D}_{k+1}\left(\boldsymbol{A}\boldsymbol{x}_k + \boldsymbol{B}_{h,k}\boldsymbol{u}_{h,k} + \boldsymbol{B}_{m,k}u_{m,k}\right) + \boldsymbol{B}_{h,k}{}^T\boldsymbol{F}_{k+1}\end{aligned} \tag{30}$$

The optimal solution $\boldsymbol{u}_{h,k}^*$ should satisfy $\frac{\partial V(\boldsymbol{x}_k)}{\partial \boldsymbol{u}_{h,k}} = 0$. Hence,

$$\boldsymbol{u}_{h,k}^* = -\left(\boldsymbol{R}_{h,k} + \boldsymbol{B}_{h,k}{}^T\boldsymbol{D}_{k+1}\boldsymbol{B}_{h,k}\right)^{-1}\boldsymbol{B}_{h,k}{}^T \left[\boldsymbol{D}_{k+1}\left(\boldsymbol{A}\boldsymbol{x}_k + \boldsymbol{B}_{m,k}\boldsymbol{u}_{m,k}\right) + \boldsymbol{F}_{k+1}\right] \tag{31}$$

By defining $\boldsymbol{H}_{h,k}$ as Eq. (19), defining $\boldsymbol{K}_{h,k}$ as Eq. (20), defining $\boldsymbol{P}_{h,k}$ as Eq. (21), and defining $\boldsymbol{S}_{h,k}$ as Eq. (22), Eq. (31) could be simplified into Eq. (18).

Substituting Eq. (18) into the vehicle dynamics model in Eq. (14):

$$\boldsymbol{x}_{k+1} = \boldsymbol{N}_k\boldsymbol{x}_k + \boldsymbol{O}_k \tag{32}$$

where

$$\boldsymbol{O}_k = \boldsymbol{B}_{h,k}\left(\boldsymbol{P}_{h,k}\boldsymbol{u}_{m,k} + \boldsymbol{S}_{h,k}\right) + \boldsymbol{B}_{m,k}\boldsymbol{u}_{m,k} \tag{33}$$

By substituting Eqs. (18) and (32) into Eq. (29), we can derive the cost-to-go function $V(\boldsymbol{x}_{k-1})$ at step $k-1$. $V(\boldsymbol{x}_{k-1})$ should also follow second-order form as the $V(\boldsymbol{x}_k)$ in Eq. (29). Hence, the coefficient matrices in Eq. (29), $\boldsymbol{D}$ and $\boldsymbol{F}$, could be derived iteratively as Eqs. (25) and (26).

The coefficient matrices in Eqs. (19)-(24) could be backward-iteratively computed from terminal step $K$ to the first step. The optimal control law is received, as shown in Eq. (18). This concludes the proof. ■

It should be noted that in ***Theorem 1***, the human's action $\boldsymbol{u}_{h,k}$ relies on the vehicle state $\boldsymbol{x}_k$ and machine's command $\boldsymbol{u}_{m,k}$. Hence, machine controller is capable of computing commands to guide human actions as the machine's expectation.

*3) Inverse Reinforcement Learning (IRL)-based driving style identification*

To adapt to various driving styles, a driving style identification method is utilized as the input of human reaction modeling, based on IRL [37]. As shown in Eq. (15), weighting factors of speed error cost $q_h^v$, gap error cost $q_h^g$, and control effort cost $R_h$ are needed to be identified person-by-person. We define $\boldsymbol{\beta}_k$ as the driving style parameter to be identified as Eq. (34).

$$\boldsymbol{\beta}_k \overset{\text{def}}{=} \left[q_{h,k}^v, q_{h,k}^g, \boldsymbol{R}_{h,k}\right]^T \tag{34}$$

The optimal control problem of human driving (13) could be transformed into Eq. (35).

$$\begin{aligned}&\min J_h = \sum_{k=1}^{k=K-1} \boldsymbol{\beta}_k{}^T\left(\frac{1}{2}\Delta v_k{}^2, \frac{1}{2}\left(g_k - g_{h,k}^{ref}\right)^2, \frac{1}{2}a_{h,k}{}^2\right)^T \\ &= \sum_{k=1}^{k=K-1} \boldsymbol{\beta}_k{}^T\boldsymbol{L}_k\left(\boldsymbol{x}_k, a_{h,k}\right) \\ &\text{s.t. vehicle dynamics: } f_h\left(\boldsymbol{x}_k, a_{h,k}, \boldsymbol{u}_{m,k}\right) = 0\end{aligned} \tag{35}$$

The driving style parameter $\boldsymbol{\beta}_k$ shall satisfy ***Proposition 1***.

***Proposition 1***: $\boldsymbol{\beta}_k$ is the solution to Eq (36).

$$\boldsymbol{\Omega}_k\mathbb{G}_k\boldsymbol{z}_1 = \boldsymbol{0} \tag{36}$$

$$\boldsymbol{z}_{k+1} = \boldsymbol{G}_k\boldsymbol{z}_k \tag{37}$$

where

$$\boldsymbol{\Omega}_k = \left(\nabla_{a_{h,k}}\boldsymbol{L}_k, \boldsymbol{B}_{h,k}{}^T\right) \tag{38}$$

$$\mathbb{G}_k = \prod_{j=1}^{j=k}\boldsymbol{G}_j \tag{39}$$

$$\boldsymbol{G}_k = \begin{pmatrix}\boldsymbol{I} & \boldsymbol{0} \\ -(\boldsymbol{A}^T)^{-1}\nabla_{x_k}\boldsymbol{L}_k & (\boldsymbol{A}^T)^{-1}\end{pmatrix} \tag{40}$$

$$\boldsymbol{z}_k = \begin{pmatrix}\boldsymbol{\beta}_k \\ \boldsymbol{\gamma}_k\end{pmatrix} \tag{41}$$

Proof:

Given the optimal solution (real-time collected human behavior), driving style parameter $\boldsymbol{\beta}_k$ should be identified to best matching the optimal solution. Based on Pontryagin's minimum principle, the solution to the optimal control problem (35) should follow the necessary conditions (42)-(43).

$$\boldsymbol{\gamma}_k = \nabla_{\boldsymbol{x}_k} H_k(\boldsymbol{x}_k, a_{h,k}, \boldsymbol{\gamma}_{k+1}, \boldsymbol{\beta}_k) \tag{42}$$

$$\nabla_{a_{h,k}} H_k(\boldsymbol{x}_k, a_{h,k}, \boldsymbol{\gamma}_{k+1}, \boldsymbol{\beta}_k) = 0 \tag{43}$$

where $H_k$ is the Hamilton's function defined as follows:

$$H_k(\boldsymbol{x}_k, a_{h,k}, \boldsymbol{\gamma}_{k+1}, \boldsymbol{\beta}_k) = \boldsymbol{\beta}_k{}^T \boldsymbol{L}_k(\boldsymbol{x}_k, a_{h,k}) + \boldsymbol{\gamma}_{k+1}{}^T(\boldsymbol{A}\boldsymbol{x}_k + \boldsymbol{B}_{h,k} a_{h,k} + \boldsymbol{B}_{m,k}\boldsymbol{u}_{m,k}) \tag{44}$$

where $\boldsymbol{\gamma}$ is the costate vector of $\boldsymbol{x}$.

Substituting the Eq. (44) into Eqs. (42) and (43):

$$\boldsymbol{\gamma}_{k+1} = \left(-(\boldsymbol{A}^T)^{-1}\nabla_{\boldsymbol{x}_k}\boldsymbol{L}_k, (\boldsymbol{A}^T)^{-1}\right)\begin{pmatrix}\boldsymbol{\beta}_k \\ \boldsymbol{\gamma}_k\end{pmatrix} \tag{45}$$

$$\left(\nabla_{a_{h,k}}\boldsymbol{L}_k, \boldsymbol{B}_{h,k}{}^T\right)\begin{pmatrix}\boldsymbol{\beta}_k \\ \boldsymbol{\gamma}_{k+1}\end{pmatrix} = \boldsymbol{0} \tag{46}$$

By defining $\boldsymbol{z}_k \overset{\text{def}}{=} \begin{pmatrix}\boldsymbol{\beta}_k \\ \boldsymbol{\gamma}_k\end{pmatrix}$, Eqs. (45) and (46) are transformed into a recursive form as Eqs. (47) and (48).

$$\boldsymbol{z}_{k+1} = \begin{pmatrix}\boldsymbol{I} & \boldsymbol{0} \\ -(\boldsymbol{A}^T)^{-1}\nabla_{\boldsymbol{x}_k}\boldsymbol{L}_k & (\boldsymbol{A}^T)^{-1}\end{pmatrix}\boldsymbol{z}_k \tag{47}$$

$$\left(\nabla_{a_{h,k}}\boldsymbol{L}_k, \boldsymbol{B}_{h,k}{}^T\right)\boldsymbol{z}_{k+1} = \boldsymbol{0} \tag{48}$$

Substituting Eq. (47) into Eq. (48) recursively, Eq. (36) is obtained. This concludes the proof. ■

Theoretically, $\boldsymbol{\beta}_k$ is exactly the solution to Eq (36). However, due to modeling error, the equality relation in Eq (36) cannot always be satisfied. Hence, ***Proposition 1*** could be relaxed to find a $\boldsymbol{\beta}_k$ with minimum modeling error. It is transformed into the least square problem as follows.

$$\min_{\boldsymbol{z}_1} \textstyle\sum_k \|\boldsymbol{\Omega}_k \mathbb{G}_k \boldsymbol{z}_1\|^2 = \min_{\boldsymbol{z}_1} \boldsymbol{z}_1{}^T \mathbb{V}\, \boldsymbol{z}_1 \tag{49}$$

$$\text{s.t. } \boldsymbol{\eta}\boldsymbol{z}_1 = 1 \tag{50}$$

where $\boldsymbol{\eta} = (1,0,0,\cdots,0)$; the constraint means that $\beta_1$ is set as 1 to maintain a same scaling during the iteration of weighting factors. $\mathbb{V}$ is semidefinite and defined as follows.

$$\mathbb{V} = \textstyle\sum_k\left((\boldsymbol{\Omega}_k\mathbb{G}_k)^T(\boldsymbol{\Omega}_k\mathbb{G}_k)\right) \tag{51}$$

By using Lagrange multiplier method, the optimal solution $\boldsymbol{z}_1{}^*$ to problem (49) could be computed by Eq. (52).

$$2\mathbb{V}\boldsymbol{z}_1{}^* + \zeta\boldsymbol{\eta}^T = \boldsymbol{0} \tag{52}$$

where $\zeta$ is the Lagrange multiplier.

Eliminating the $\zeta$ by dividing $\boldsymbol{z}_1$ as $\boldsymbol{z}_1 = \begin{bmatrix}\beta_1 \\ \underline{\boldsymbol{z}}_1\end{bmatrix}$, following equation is obtained.

$$\mathbb{V}\boldsymbol{z}_1{}^* = \begin{pmatrix}\overline{\mathbb{V}} & \boldsymbol{V}^T \\ \boldsymbol{V} & \underline{\mathbb{V}}\end{pmatrix}\begin{bmatrix}1 \\ \underline{\boldsymbol{z}}_1{}^*\end{bmatrix} = -\frac{1}{2}\zeta\boldsymbol{\eta}^T \tag{53}$$

where $\mathbb{V}$ is decomposed into four modules; $\boldsymbol{V}$ is a column vector; $\overline{\mathbb{V}}$ and $\underline{\mathbb{V}}$ are symmetric matrices.

Eq. (53) gives the iterative law as follows:

$$\underline{\boldsymbol{z}}_1{}^* = -\underline{\mathbb{V}}^{-1}\boldsymbol{V} \tag{54}$$

Molloy et al. have proven that $\boldsymbol{z}_1{}^*$ would converge with the increase of the size of $\mathbb{V}$ (more data are collected) [37]. According to Eq. (41), the first three elements of $\boldsymbol{z}_1{}^*$ would be the optimal driving style parameter $\boldsymbol{\beta}_1{}^*$. In online implementation, the IRL-based driving style identification algorithm is summarized as ***Algorithm 1***.

***Algorithm 1***: IRL-based driving style identification

**Input** $\boldsymbol{\beta}^0$ and modeling error tolerance $\varepsilon$
**While** $\|\boldsymbol{\beta}^* - \boldsymbol{\beta}^0\| > \varepsilon$ **do**:
  Updating from last iteration: $\boldsymbol{\beta}^0 = \boldsymbol{\beta}^*$
  Collecting human driving data accumulatively
  Calculating $\boldsymbol{\beta}^*$ based on Eq. (54)
**Output** $\boldsymbol{\beta}^*$

### *E. Machine controller*

The machine controller is developed to ensure following stability during the takeover maneuver. The controller is formulated into a GMPC problem within the Stachelberg competition framework. The human reaction function is substituted into the human-machine shared system dynamics model so that the optimization of the machine's command takes full account of human actions accordingly.

#### *1) GMPC based controller*

The machine controller could be formulated into a GMPC problem as follows. It aims at finding the optimal control series $\boldsymbol{u}_{m,k}\big|_{k=1}^{k=K}$ to enhance the following stability.

$$\min_{\boldsymbol{u}_{m,k}|_{k=1}^{k=K}}(J_m) = \textstyle\sum_{k=1}^{k=K-1}\Phi_{m,k}(\boldsymbol{x}_k, \boldsymbol{u}_{m,k}, \boldsymbol{x}_{m,k}^{ref}) + \Psi_{m,K}(\boldsymbol{x}_k, \boldsymbol{x}_{m,k}^{ref}) \tag{55}$$

$$\text{s.t. vehicle dynamics: } f_m(\boldsymbol{x}_k, \boldsymbol{u}_{m,k}) = 0 \tag{56}$$

$$\text{State constraint: } \boldsymbol{x}_{min,k} \le \boldsymbol{x}_k \le \boldsymbol{x}_{max,k} \tag{57}$$

$$\text{Control constraint: } \boldsymbol{u}_{min,k} \le \boldsymbol{u}_{m,k} \le \boldsymbol{u}_{max,k} \tag{58}$$

where $\Phi_{m,k}$ is running cost of machine driving; $\Psi_{m,K}$ is the terminal cost of machine driving; $f_m$ is the vehicle dynamics. The constraint on human control is not presented, since it could be transformed into the constraints on state and machine control based on Eq. (18). State constraint (57) is designed to regulate the following gap greater than the minimum safe following gap. Control constraint (58) is designed to restrict the acceleration command within vehicle's execution capability. The violation of constraints shall be allowed to satisfy emergency conditions. Fallback strategies [38] and dynamic relaxation of non-safety-critical constraints [39] could be adopted.

The three functions $\Phi_{m,k}$, $\Psi_{m,K}$, and $f_m$ are formulated as follows.

$$\Phi_{m,k} = \frac{1}{2}(\boldsymbol{x}_k - \boldsymbol{x}_{m,k}^{ref})^T\boldsymbol{Q}_m(\boldsymbol{x}_k - \boldsymbol{x}_{m,k}^{ref}) + \frac{1}{2}\boldsymbol{u}_{m,k}{}^T R_m \boldsymbol{u}_{m,k} \tag{59}$$

$$\Psi_{m,K} = \frac{1}{2}(\boldsymbol{x}_K - \boldsymbol{x}_{m,K}^{ref})^T\boldsymbol{Q}_m(\boldsymbol{x}_K - \boldsymbol{x}_{m,K}^{ref}) \tag{60}$$

$$f_m = \boldsymbol{x}_{k+1} - \boldsymbol{A}\boldsymbol{x}_k + \boldsymbol{B}_{h,k}(\boldsymbol{K}_{h,k}\boldsymbol{x}_k + \boldsymbol{P}_{h,k}\boldsymbol{u}_{m,k} + \boldsymbol{S}_{h,k}) + \boldsymbol{B}_{m,k}\boldsymbol{u}_{m,k} + \boldsymbol{C}a_k^p \tag{61}$$

where $\boldsymbol{Q}_m = \text{diag}[q_m^v, q_m^g]$. $q_m^v$, $q_m^g$, and $R_m$ are weighting factors of speed error cost, gap error cost, and control effort cost, respectively. $q_m^v$ is set as a positive value and $q_m^g$ is typically a relatively small positive value, since machine's objective is to ensure following stability at all following gaps. The vehicle dynamics model Eq. (61) is obtained from Eq. (12) after substituting human's reaction function in Eq. (18).

*2) Problem transformation*

The GMPC problem in Eq. (55) consists of a time-varying system dynamics model, as shown in Eq. (61). Furthermore, the non-aftereffect property is not applicable anymore, since the coefficient matrix $\boldsymbol{S}_{h,k}$ consists of an iterative relation of $\boldsymbol{u}_{m,i}\{\forall i \in [k+1, K]\}$ . Hence, the GMPC problem is transformed into a Quadratic Programming (QP) problem for solving.

Defining optimization variables by containing all variables from $k=1$ to $k=K$ as follows:

$$\boldsymbol{U}_h \stackrel{\text{def}}{=} \left[\boldsymbol{u}_{h,1}, \boldsymbol{u}_{h,2}, \cdots, \boldsymbol{u}_{h,k}, \cdots, \boldsymbol{u}_{h,K-1}\right]^T \tag{62}$$

$$\boldsymbol{U}_m \stackrel{\text{def}}{=} \left[\boldsymbol{u}_{m,1}, \boldsymbol{u}_{m,2}, \cdots, \boldsymbol{u}_{m,k}, \cdots, \boldsymbol{u}_{m,K-1}\right]^T \tag{63}$$

$$\boldsymbol{X} \stackrel{\text{def}}{=} [\boldsymbol{x}_1, \boldsymbol{x}_2, \cdots, \boldsymbol{x}_k, \cdots, \boldsymbol{x}_K]^T \tag{64}$$

The human reaction function (18) in ***Theorem 1*** could be converted into an integrated form as Eq. (65).

$$\boldsymbol{U}_h = \boldsymbol{\mathcal{K}}_h \boldsymbol{X} + (\boldsymbol{\mathcal{P}}_h + \boldsymbol{\mathcal{T}}_h \boldsymbol{\mathcal{S}}_{h1})\boldsymbol{U}_m + \boldsymbol{\mathcal{T}}_h \boldsymbol{\mathcal{S}}_{h2} \tag{65}$$

where

$$\boldsymbol{\mathcal{K}}_h \stackrel{\text{def}}{=} \begin{bmatrix} \boldsymbol{K}_{h,1} & \boldsymbol{0} & \cdots & \boldsymbol{0} & \boldsymbol{0} \\ \boldsymbol{0} & \boldsymbol{K}_{h,2} & \cdots & \boldsymbol{0} & \boldsymbol{0} \\ \vdots & \vdots & \ddots & \boldsymbol{0} & \vdots \\ \boldsymbol{0} & \boldsymbol{0} & \cdots & \boldsymbol{K}_{h,K-1} & \boldsymbol{0} \end{bmatrix} \tag{66}$$

$$\boldsymbol{\mathcal{P}}_h \stackrel{\text{def}}{=} \begin{bmatrix} \boldsymbol{P}_{h,1} & 0 & \cdots & 0 \\ 0 & \boldsymbol{P}_{h,2} & \cdots & 0 \\ \vdots & \vdots & \ddots & 0 \\ 0 & 0 & \cdots & \boldsymbol{P}_{h,K-1} \end{bmatrix} \tag{67}$$

$$\boldsymbol{\mathcal{T}}_h \stackrel{\text{def}}{=} \begin{bmatrix} -\boldsymbol{H}_{h,1}\boldsymbol{B}_{h,1}{}^T & \boldsymbol{0} & \cdots & \boldsymbol{0} \\ \boldsymbol{0} & -\boldsymbol{H}_{h,2}\boldsymbol{B}_{h,2}{}^T & \cdots & \boldsymbol{0} \\ \vdots & \vdots & \ddots & \boldsymbol{0} \\ \boldsymbol{0} & \boldsymbol{0} & \cdots & -\boldsymbol{H}_{h,K-1}\boldsymbol{B}_{h,K-1}{}^T \end{bmatrix} \tag{68}$$

$$\boldsymbol{\mathcal{S}}_{h1} \stackrel{\text{def}}{=} \begin{bmatrix} \boldsymbol{0}(\boldsymbol{M}_{h,2} + \boldsymbol{N}_2{}^T)\boldsymbol{D}_3\boldsymbol{B}_{m,2} \cdots & \prod_{w=2}^{c}\left((\boldsymbol{M}_{h,w} + \boldsymbol{N}_w{}^T)\right)\boldsymbol{D}_{c+1}\boldsymbol{B}_{m,c} & \cdots & \prod_{w=2}^{K-1}(\boldsymbol{M}_{h,w} + \boldsymbol{N}_w{}^T)\boldsymbol{D}_K\boldsymbol{B}_{m,K-1} \\ \boldsymbol{0} \quad \boldsymbol{0} \quad \ddots & \vdots & \cdots & \vdots \\ \boldsymbol{0} \quad \boldsymbol{0} \quad \ddots & \prod_{w=r+1}^{c}\left((\boldsymbol{M}_{h,w} + \boldsymbol{N}_w{}^T)\right)\boldsymbol{D}_{c+1}\boldsymbol{B}_{m,c} & \cdots & \prod_{w=r+1}^{K-1}\left((\boldsymbol{M}_{h,w} + \boldsymbol{N}_w{}^T)\right)\boldsymbol{D}_K\boldsymbol{B}_{m,K-1} \\ \vdots \quad \vdots \quad \vdots & \ddots & \cdots & \vdots \\ \boldsymbol{0} \quad \boldsymbol{0} \quad \boldsymbol{0} & \boldsymbol{0} & \boldsymbol{0} & (\boldsymbol{M}_{h,K-1} + \boldsymbol{N}_{K-1}{}^T)\boldsymbol{D}_K\boldsymbol{B}_{m,K-1} \\ \boldsymbol{0} \quad \boldsymbol{0} \quad \boldsymbol{0} & \boldsymbol{0} & \cdots & \boldsymbol{0} \end{bmatrix} \tag{69}$$

$$\boldsymbol{\mathcal{S}}_{h2} \stackrel{\text{def}}{=} \begin{bmatrix} -\sum_{j=1}^{K-1}\left(\prod_{w=2}^{j}\left((\boldsymbol{M}_{h,w} + \boldsymbol{N}_w{}^T)\right)\boldsymbol{Q}_h\boldsymbol{x}_{h,j+1}^{ref}\right) \\ \vdots \\ -\sum_{j=r}^{K-1}\left(\prod_{w=r+1}^{j}\left((\boldsymbol{M}_{h,w} + \boldsymbol{N}_w{}^T)\right)\boldsymbol{Q}_h\boldsymbol{x}_{h,j+1}^{ref}\right) \\ \vdots \\ -\boldsymbol{Q}_h\boldsymbol{x}_{h,K}^{ref} \end{bmatrix} \tag{70}$$

Finally, the GMPC problem in Eq. (55) could be transformed into a QP form as follows.

$$\min_{\boldsymbol{U}_m}(J_m) = \frac{1}{2}\begin{pmatrix}\boldsymbol{X} \\ \boldsymbol{U}_m\end{pmatrix}^T \mathbb{D}_m \begin{pmatrix}\boldsymbol{X} \\ \boldsymbol{U}_m\end{pmatrix} + \begin{pmatrix}\boldsymbol{X} \\ \boldsymbol{U}_m\end{pmatrix}^T \mathbb{F}_m \tag{71}$$

$$\text{s.t.} \quad (\boldsymbol{I} - \boldsymbol{\mathcal{A}}_m - \boldsymbol{\mathcal{B}}_h\boldsymbol{\mathcal{K}}_h \quad -\boldsymbol{\mathcal{B}}_h\boldsymbol{\mathcal{P}}_h - \boldsymbol{\mathcal{B}}_h\boldsymbol{\mathcal{T}}_h\boldsymbol{\mathcal{S}}_{h1} - \boldsymbol{\mathcal{B}}_m)\begin{pmatrix}\boldsymbol{X} \\ \boldsymbol{U}_m\end{pmatrix} = \boldsymbol{\mathcal{B}}_h\boldsymbol{\mathcal{T}}_h\boldsymbol{\mathcal{S}}_{h2} + \boldsymbol{\mathcal{C}}_m \tag{72}$$

where

$$\boldsymbol{X}_m^{ref} \stackrel{\text{def}}{=} \left[\boldsymbol{x}_{m,1}^{ref}, \boldsymbol{x}_{m,2}^{ref}, \cdots, \boldsymbol{x}_{m,k}^{ref}, \cdots, \boldsymbol{x}_{m,K}^{ref}\right]^T \tag{73}$$

$$\boldsymbol{\mathcal{Q}}_m \stackrel{\text{def}}{=} \begin{bmatrix} \boldsymbol{Q}_m & \boldsymbol{0} & \cdots & \boldsymbol{0} \\ \boldsymbol{0} & \boldsymbol{Q}_m & \cdots & \boldsymbol{0} \\ \vdots & \vdots & \ddots & \boldsymbol{0} \\ \boldsymbol{0} & \boldsymbol{0} & \cdots & \boldsymbol{Q}_m \end{bmatrix} \tag{74}$$

$$\boldsymbol{\mathcal{R}}_m \stackrel{\text{def}}{=} \begin{bmatrix} R_m & 0 & \cdots & 0 \\ 0 & R_m & \cdots & 0 \\ \vdots & \vdots & \ddots & 0 \\ 0 & 0 & \cdots & R_m \end{bmatrix} \tag{75}$$

$$\boldsymbol{\mathcal{A}}_m \stackrel{\text{def}}{=} \begin{bmatrix} \boldsymbol{0} & \boldsymbol{0} & \cdots & \boldsymbol{0} & \boldsymbol{0} \\ \boldsymbol{A} & \boldsymbol{0} & \cdots & \boldsymbol{0} & \boldsymbol{0} \\ \boldsymbol{0} & \boldsymbol{A} & \ddots & \boldsymbol{0} & \boldsymbol{0} \\ \vdots & \vdots & \ddots & \vdots & \vdots \\ \boldsymbol{0} & \boldsymbol{0} & \cdots & \boldsymbol{A} & \boldsymbol{0} \end{bmatrix} \tag{76}$$

$$\boldsymbol{\mathcal{B}}_h \stackrel{\text{def}}{=} \begin{bmatrix} \boldsymbol{0} & \boldsymbol{0} & \cdots & \boldsymbol{0} \\ \boldsymbol{B}_{h,1} & \boldsymbol{0} & \cdots & \boldsymbol{0} \\ \boldsymbol{0} & \boldsymbol{B}_{h,2} & \ddots & \boldsymbol{0} \\ \vdots & \vdots & \ddots & \vdots \\ \boldsymbol{0} & \boldsymbol{0} & \cdots & \boldsymbol{B}_{h,K-1} \end{bmatrix} \tag{77}$$

$$\boldsymbol{\mathcal{B}}_m \stackrel{\text{def}}{=} \begin{bmatrix} \boldsymbol{0} & \boldsymbol{0} & \cdots & \boldsymbol{0} \\ \boldsymbol{B}_{m,1} & \boldsymbol{0} & \cdots & \boldsymbol{0} \\ \boldsymbol{0} & \boldsymbol{B}_{m,2} & \ddots & \boldsymbol{0} \\ \vdots & \vdots & \ddots & \vdots \\ \boldsymbol{0} & \boldsymbol{0} & \cdots & \boldsymbol{B}_{m,K-1} \end{bmatrix} \tag{78}$$

$$\boldsymbol{\mathcal{C}}_m \stackrel{\text{def}}{=} \begin{bmatrix} \boldsymbol{x}_1 \\ \boldsymbol{0} \\ \vdots \\ \boldsymbol{0} \end{bmatrix} + \begin{bmatrix} \boldsymbol{0} & \boldsymbol{0} & \cdots & \boldsymbol{0} \\ \boldsymbol{C} & \boldsymbol{0} & \cdots & \boldsymbol{0} \\ \boldsymbol{0} & \boldsymbol{C} & \ddots & \boldsymbol{0} \\ \vdots & \vdots & \ddots & \vdots \\ \boldsymbol{0} & \boldsymbol{0} & \cdots & \boldsymbol{C} \end{bmatrix} \boldsymbol{a}^p \tag{79}$$

$$\mathbb{D}_m \stackrel{\text{def}}{=} \begin{bmatrix} \boldsymbol{\mathcal{Q}}_m & \boldsymbol{0} \\ \boldsymbol{0} & \boldsymbol{\mathcal{R}}_m \end{bmatrix} \tag{80}$$

$$\mathbb{F}_m \stackrel{\text{def}}{=} \begin{bmatrix} -\boldsymbol{\mathcal{Q}}_m\boldsymbol{X}_m^{ref} \\ \boldsymbol{0} \end{bmatrix} \tag{81}$$

where $\boldsymbol{a}^p \stackrel{\text{def}}{=} \left[a_1^p, a_2^p, \cdots, a_{K-1}^p\right]^T$.

*3) Optimal solution*

The QP problem (71) could be solved by the iterative Lagrange multiplier method. Since stable following and less control efforts are included in the objective function (55), the constraints in Eqs. (57) and (58) are hard to be activated. Hence, they are not considered in the solution. We use an action safeguard to check if the solution is within the constraint [40]. If constraints are violated, the nearest feasible solution on the constraint boundary is adopted.

***Theorem 2***: The optimal solution of problem (71) is as follows.

$$\begin{pmatrix} \boldsymbol{X} \\ \boldsymbol{U}_m \\ \boldsymbol{\lambda}^T \end{pmatrix} = \begin{bmatrix} \mathbb{D}_m & \mathbb{W}_m{}^T \\ \mathbb{W}_m & \mathbf{0} \end{bmatrix}^{-1} \begin{bmatrix} -\mathbb{F}_m \\ \mathbb{Z}_m \end{bmatrix} \tag{82}$$

$$\mathbb{W}_m \stackrel{\text{def}}{=} (\boldsymbol{I} - \boldsymbol{\mathcal{A}}_m - \boldsymbol{\mathcal{B}}_h \boldsymbol{\mathcal{K}}_h \quad -\boldsymbol{\mathcal{B}}_h \boldsymbol{\mathcal{P}}_h - \boldsymbol{\mathcal{B}}_h \boldsymbol{\mathcal{T}}_h \boldsymbol{\mathcal{S}}_{h1} - \boldsymbol{\mathcal{B}}_m) \tag{83}$$

$$\mathbb{Z}_m \stackrel{\text{def}}{=} \boldsymbol{\mathcal{B}}_h \boldsymbol{\mathcal{T}}_h \boldsymbol{\mathcal{S}}_{h2} + \boldsymbol{\mathcal{C}}_m \tag{84}$$

where $\boldsymbol{\lambda}$ is the Lagrange multiplier.

Proof:

Introducing Lagrange multiplier $\boldsymbol{\lambda}$ and defining Lagrange function $\mathbb{L}$ as:

$$\mathbb{L} \stackrel{\text{def}}{=} \frac{1}{2} \begin{pmatrix} \boldsymbol{X} \\ \boldsymbol{U}_m \end{pmatrix}^T \mathbb{D}_m \begin{pmatrix} \boldsymbol{X} \\ \boldsymbol{U}_m \end{pmatrix} + \begin{pmatrix} \boldsymbol{X} \\ \boldsymbol{U}_m \end{pmatrix}^T \mathbb{F}_m + \boldsymbol{\lambda} \left[ \mathbb{W}_m \begin{pmatrix} \boldsymbol{X} \\ \boldsymbol{U}_m \end{pmatrix} - \mathbb{Z}_m \right] \tag{85}$$

The optimal solution $\begin{pmatrix} \boldsymbol{X}^* \\ \boldsymbol{U}_m{}^* \end{pmatrix}$ is obtained by computing the derivative of $\mathbb{L}$:

$$\frac{\partial \mathbb{L}}{\partial \begin{pmatrix} \boldsymbol{X} \\ \boldsymbol{U}_m \end{pmatrix}} = \mathbb{D}_m \begin{pmatrix} \boldsymbol{X} \\ \boldsymbol{U}_m \end{pmatrix} + \mathbb{F}_m + \mathbb{W}_m{}^T \boldsymbol{\lambda}^T = \mathbf{0} \tag{86}$$

$$\frac{\partial \mathbb{L}}{\partial \lambda} = \mathbb{W}_m \begin{pmatrix} \boldsymbol{X} \\ \boldsymbol{U}_m \end{pmatrix} - \mathbb{Z}_m = \mathbf{0} \tag{87}$$

Solving the Eqs. (86) and (87), the optimal solution could be computed into Eq. (82). This concludes the proof. ■

## III. Evaluation

Simulation is conducted on the Matlab platform. The proposed human-machine shared CACC controller is evaluated from the following perspectives: i) the function of smooth takeover; ii) platoon sting stability; iii) perceived safety in regard to traffic oscillations; iv) actual safety in regard to hard brakes; v) influence on upstream traffic.

### *A. Controller types*

There are two types of CACC controllers to be tested:

- **The proposed human-machine shared CACC controller**: This controller is capable of allocating control authority between human and machine. Different allocation factors could be selected. The machine could generate proactive control commands with the consideration of the preceding vehicle's future commands.
- **Baseline human driver**: Human drivers directly take over the control authority from the machine. Human actions are delayed without acquiring the preceding vehicle's future actions.

### *B. Test scenarios*

The proposed CACC controller is tested on the highway environment. The CACC platoon is following a preceding HV. CACC takeover is needed from time to time. For example, human drivers may be asked to take over the CACC platoon in order to get off the highway. Three cases are designed without the consideration of communication and control delay.

- **Case 1**: The preceding HV is cruising at a constant speed of 10 m/s. The proposed CACC controller adopts three types of takeover strategies, including i) human's direct takeover (100% of authority is given to human); ii) constant allocation (30% of authority is given to human), and iii) linear gradient from machine to human. Different takeover durations are set for the third strategy.
- **Case 2**: The preceding HV imposes a speed oscillation influence on the CACC platoon. The preceding HV's speed oscillation follows a sinusoidal function.
- **Case 3**: The preceding HV conducts a hard brake.

### *C. Measures of effectiveness*

Measures of Effectiveness (MOEs) adopted are as follows.

- **Function validation**: The function of smoothly taking over a CACC system is validated by trajectories of the following gap, speed difference, and acceleration. This MOE is tested in the scenario of Case 1.
- **String stability**: This MOE is evaluated by vehicles' following gap trajectories and gap range. Moreover, the ODD to ensure string stability is quantified by the oscillation propagation rate within the platoon $\Theta_{n,i,i-1}$, calculated as follows. This MOE is tested in the scenario of Case 2.

$$\Theta_{n,i,i-1} \stackrel{\text{def}}{=} E\left(\frac{\|g^i\|_2}{\|g^{i-1}\|_2}\right), \ \forall i \in [2, n] \tag{88}$$

where $n$ is the number of CAVs within a platoon; $\Theta_{n,i,i-1}$ is the oscillation propagation rate between the CAV $i$ and its preceding CAV $i-1$

- **Human driving style identification accuracy**: The human driving style identification accuracy is evaluated by modeling accuracy of driving style parameter $\boldsymbol{\beta}$.
- **Perceived safety**: This MOE is evaluated by acceleration oscillation range [41-44]. It is tested in the scenario of Case 2.
- **Actual safety**: This MOE is evaluated by vehicles' following gap trajectories and gap range [45-47]. This MOE is tested in the scenario of Case 3.
- **Traffic influence**: CACC's influence on upstream traffic is evaluated by vehicle trajectories and quantified by speed oscillation ranges. The influence range is identified. This MOE is tested in the scenario of Case 2.
- **Computational efficiency**: This MOE is quantified by the computing time of the proposed controller.

### *D. Results*

Results demonstrate that the proposed CACC takeover controller is capable of i) dynamically allocating control authority between machine and human to ensure a smooth takeover; ii) ensure string stability in the condition that the platoon has less than 6 CAVs and human control authority is less than 40%; iii) enhancing perceived safety and actual safety by machine's intervention; iv) reducing influence on upstream traffic.

#### *1) Function validation*

The function of smoothly taking over a CACC system is demonstrated by following gap and speed difference trajectories, as shown in Fig. 3. Three types of takeover strategies are tested, including human's direct takeover (Fig. 3-I), 30% of authority to human (Fig. 3-II), and a linear gradient of authority (Fig. 3-III). Results show that the involvement of the machine would obtain a smoother CACC takeover maneuver, as illustrated in Fig. 3 (II) and (III). Compared to the human driver's direct takeover, the proposed method reduces speed oscillation magnitude by extending the duration of the takeover. It indicates that the proposed controller is good at making a tradeoff between space and time.

The function of human-machine shared control is verified by vehicles' command trajectories. The third CAV's acceleration and its compositions from human and machine are selected for illustrations, as shown in Fig. 4. Compared to human's direct takeover in Fig. 4 (I), allocating 70% of authority to machine (Fig. 4-II) could avoid hard brakes at the beginning of takeover. Furthermore, a linear gradient authority allocation strategy can better smooth the takeover trajectory, as shown in Fig. 4 (III). In this case, with the decrease of the machine's authority and increase of human's authority, the vehicle's final acceleration command exhibits an increasing dominance by the human driver. It enables CACC to gradually hand over the control authority from machine to human. At the beginning of the takeover, the machine has greater control authority. Final acceleration has a larger composition from the machine, as illustrated by the green bar in Fig. 4. This avoids human stress behaving at the beginning of the takeover, where there is no more hard braking as shown by the red solid line in Fig. 4 (III). When the human driver gets into its stride typically within 10 seconds, more authority would be allocated to human to gradually accomplish the CACC takeover maneuver.

A sensitivity analysis was conducted on control authority takeover duration, as illustrated in Fig. 5. The results indicate that the system allows for a gradual handover of control from the machine to the human, completing the takeover process successfully. However, as the takeover duration extends from 5 seconds to 30 seconds, the total time required to stabilize the platoon increases from 28.1 seconds to 36.5 seconds. For every additional 5 seconds in takeover duration, the system requires an additional 1.68 seconds to stabilize the platoon.

*2) Human driving style identification accuracy*

The proposed IRL-based driving style identification method is confirmed to ensure the identification accuracy by 91.6%. As shown in Fig. 6, $\beta_2$ is set as 0.5 and $\beta_3$ is set as 2.5. The identified $\beta_2$ finally converges to 0.45 within 6 iteration steps. The identified $\beta_3$ finally converges to 2.32 within 5 iteration steps.

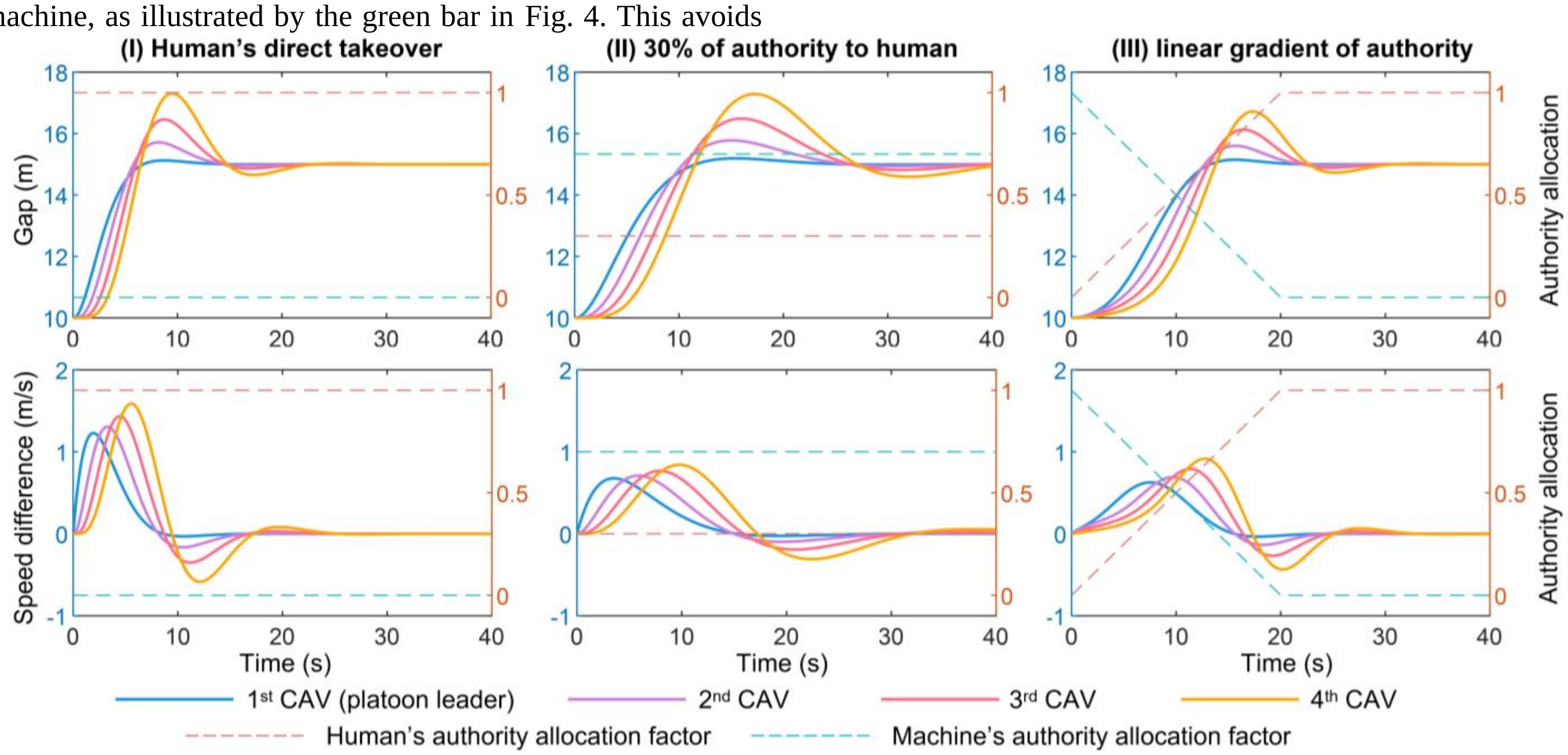

Fig. 3. Trajectories of a CACC takeover maneuver for the proposed method.

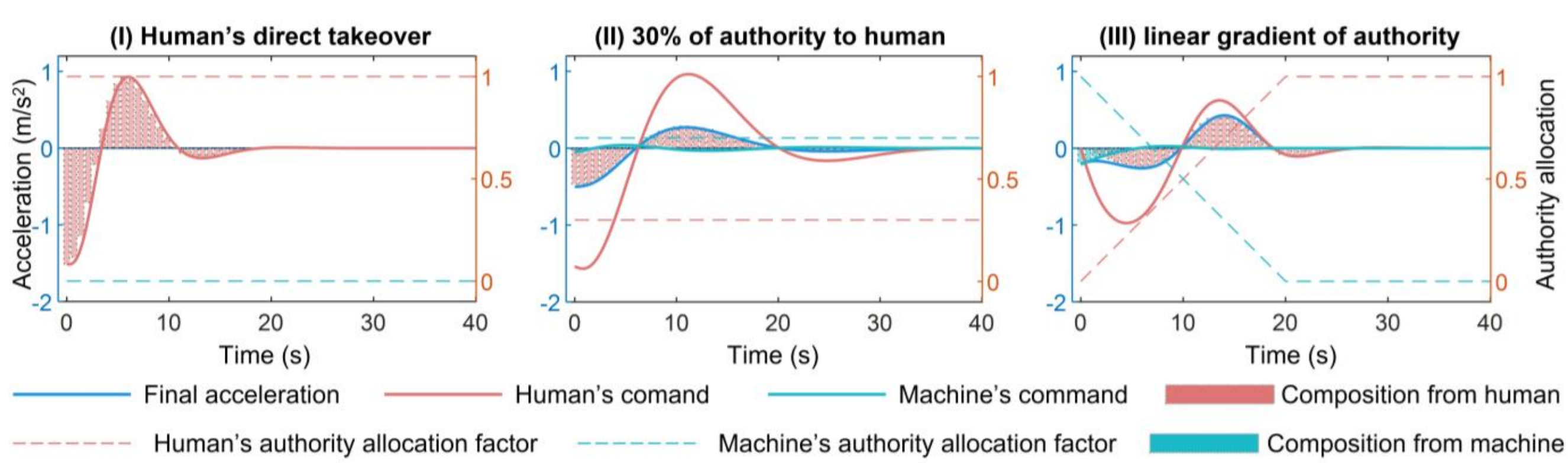

Fig. 4. Human-machine control authority allocation process.

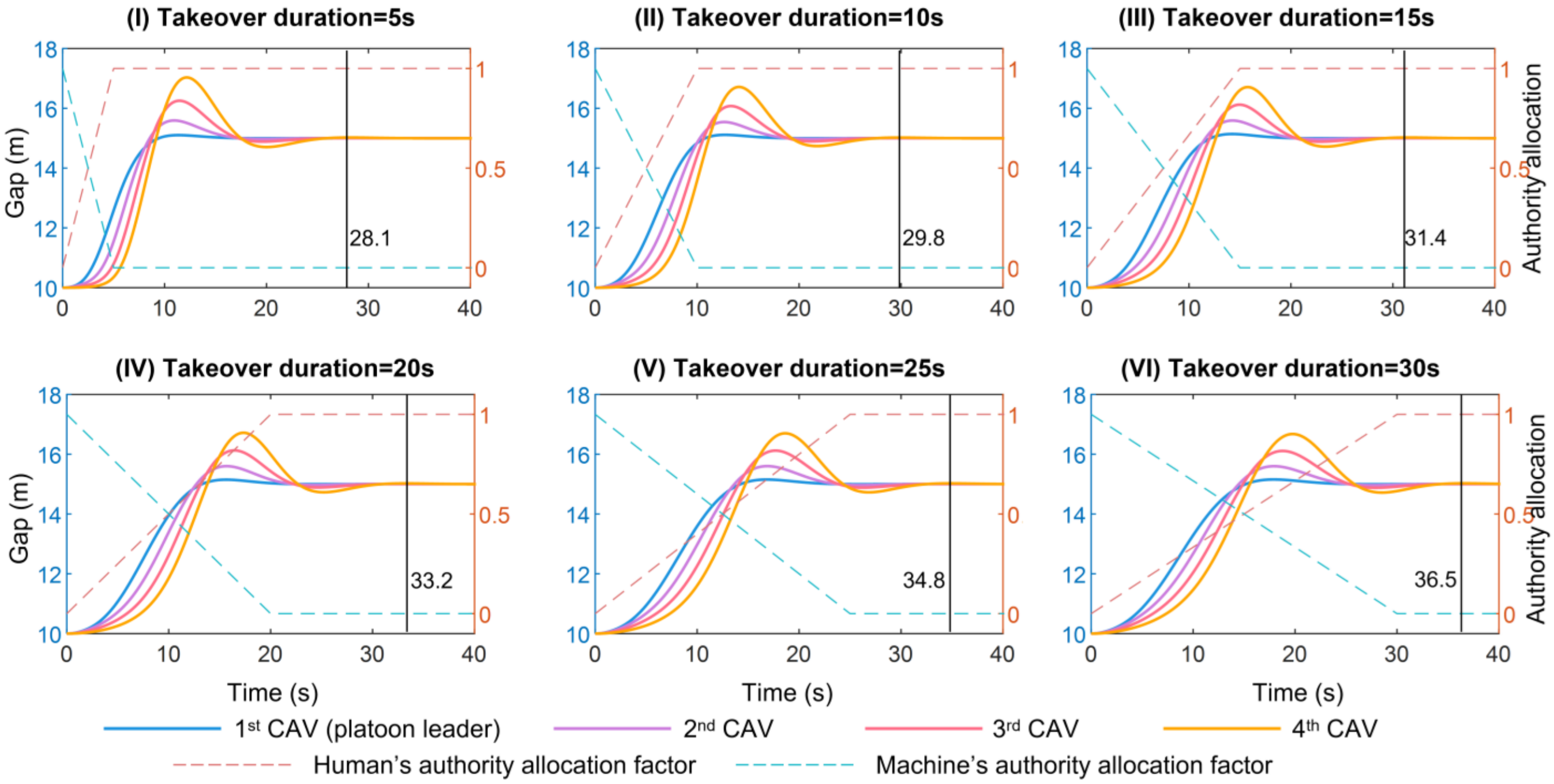

Fig. 5. Sensitivity test on control authority takeover duration.

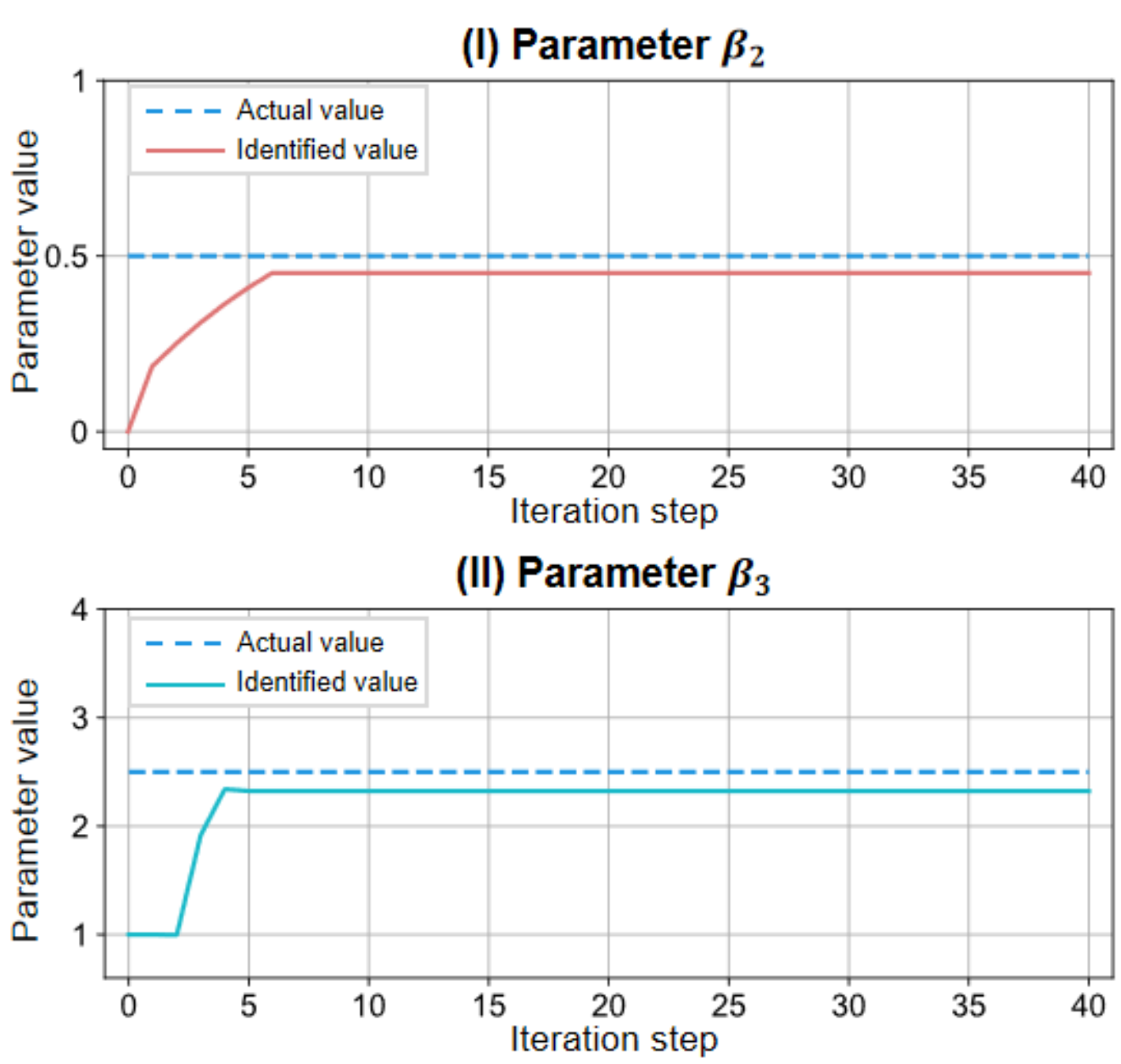

Fig. 6. Driving style identification results.

### 3) String stability

A sensitivity analysis is conducted on string stability in terms of various authority allocation factors for the proposed CACC controller. The following gap trajectories are illustrated in Fig. 7. It shows that the following gap oscillations are reduced when propagating along the platoon string, when the CACC platoon is entirely controlled by machine, as shown in Fig. 7 (I), or 20% of control authorities are allocated to human, as shown in Fig. 7 (II). This behavior is consistent with prior real-world studies on CACC performance [48, 49]. When more than 40% of control authorities are given to human, as shown in Fig. 7 (III) and (IV), the following gap oscillations are amplified by the following vehicles. System responses begin to resemble that of ACC [48]. In these cases, string stability is not ensured.

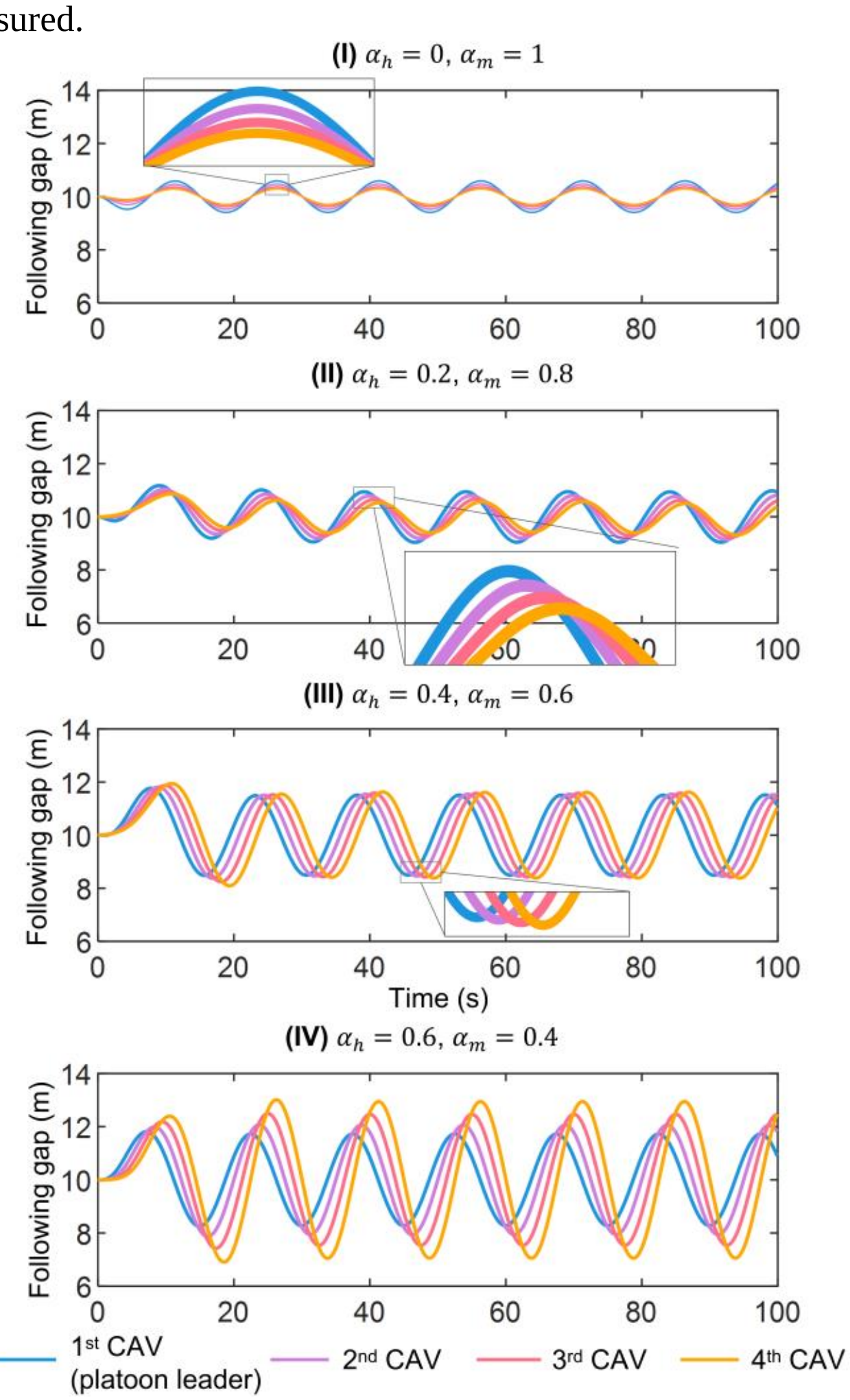

Fig. 7. Sensitivity analysis on following gap trajectories in terms of authority allocation factors.

The ODD of ensuring CACC's string stability is quantified in regard to various control authority allocation factors and the length of platoon, as shown in Fig. 8. Results show that string stability performance deteriorates with the increase of allocation factor and platoon length. When there are only two CAVs in a platoon, string stability is ensured when there are less than 45.1% of control authorities are allocated to human. However, when there are 10 CAVs in a platoon, it is required to allocate less than 28.6% of control authorities to human to ensure string stability. Furthermore, an amplified oscillation propagation is observed when the number of CAVs in a platoon increases from 6 to 7. Hence, in actual implementation, the platoon length is suggested to be restricted under 6 CAVs to ensure string stability with the human control authority allocation factor less than 40%.

*4) Perceived safety*

The perceived safety is evaluated by the acceleration range in three different authority allocation factors, as illustrated in Fig. 9. Results show that 30% of the machine's control authority (Fig. 9-II) would enhance perceived safety by 53.23%, compared to fully human's control (Fig. 9-IV). When string stability is ensured, as shown in Fig. 9 (I) and (II), the acceleration range is reduced along the platoon string. This indicates that all CAVs would be enabled with perceived safety, no matter how long the CACC platoon is. However, with the decrease of the machine's authority (from Fig. 9-I to IV), the acceleration range of followers increases. Under full human's control ($\alpha_h = 1$), as shown in Fig. 9 (IV), all following CAVs have greater acceleration range compared to the preceding HV. Moreover, the acceleration range is amplified along the platoon string. It indicates that perceived safety would be reduced along platoon string under the control of human drivers.

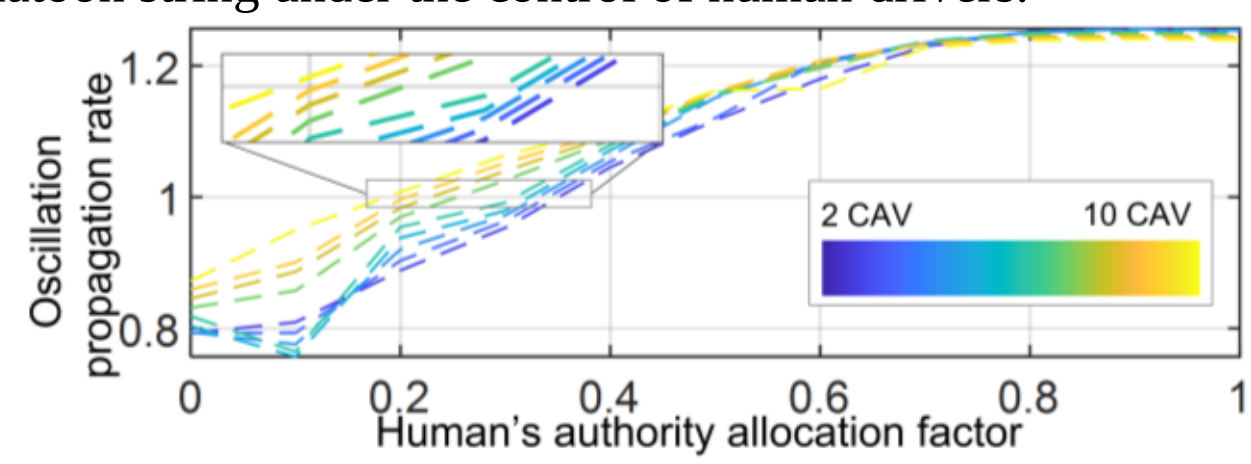

Fig. 8. String-stable ODD in regard to allocation factor and platoon length.

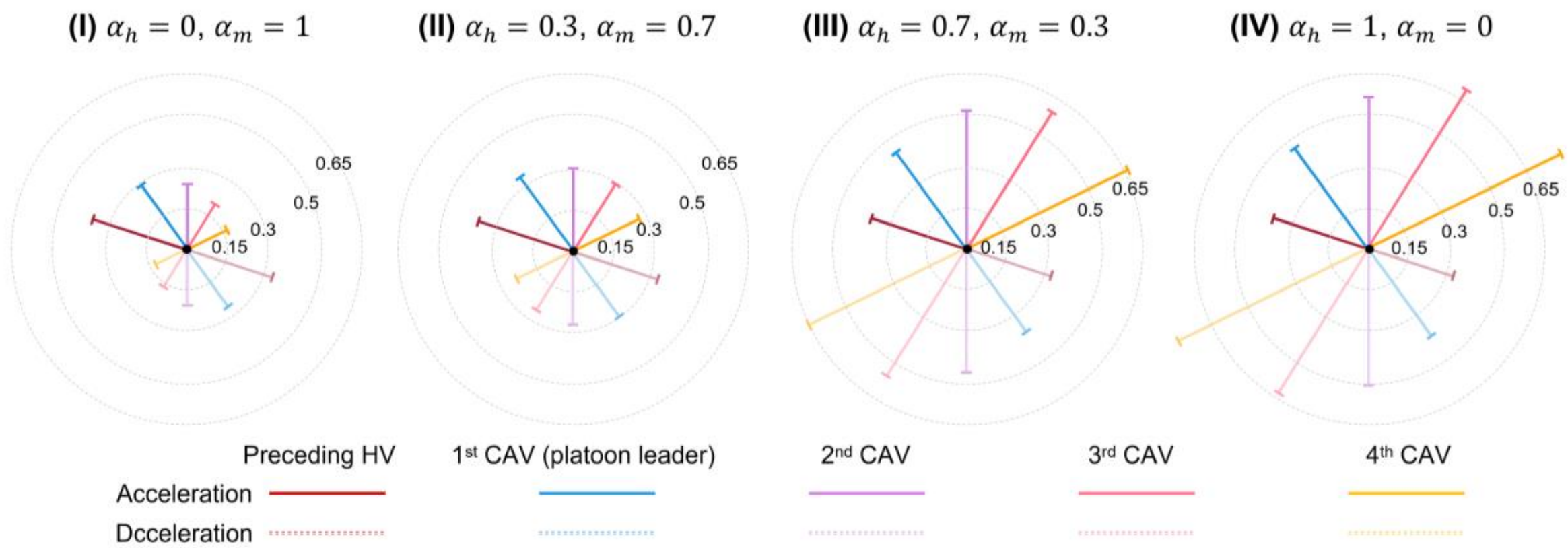

Fig. 9. Acceleration range in various authority allocation factors.

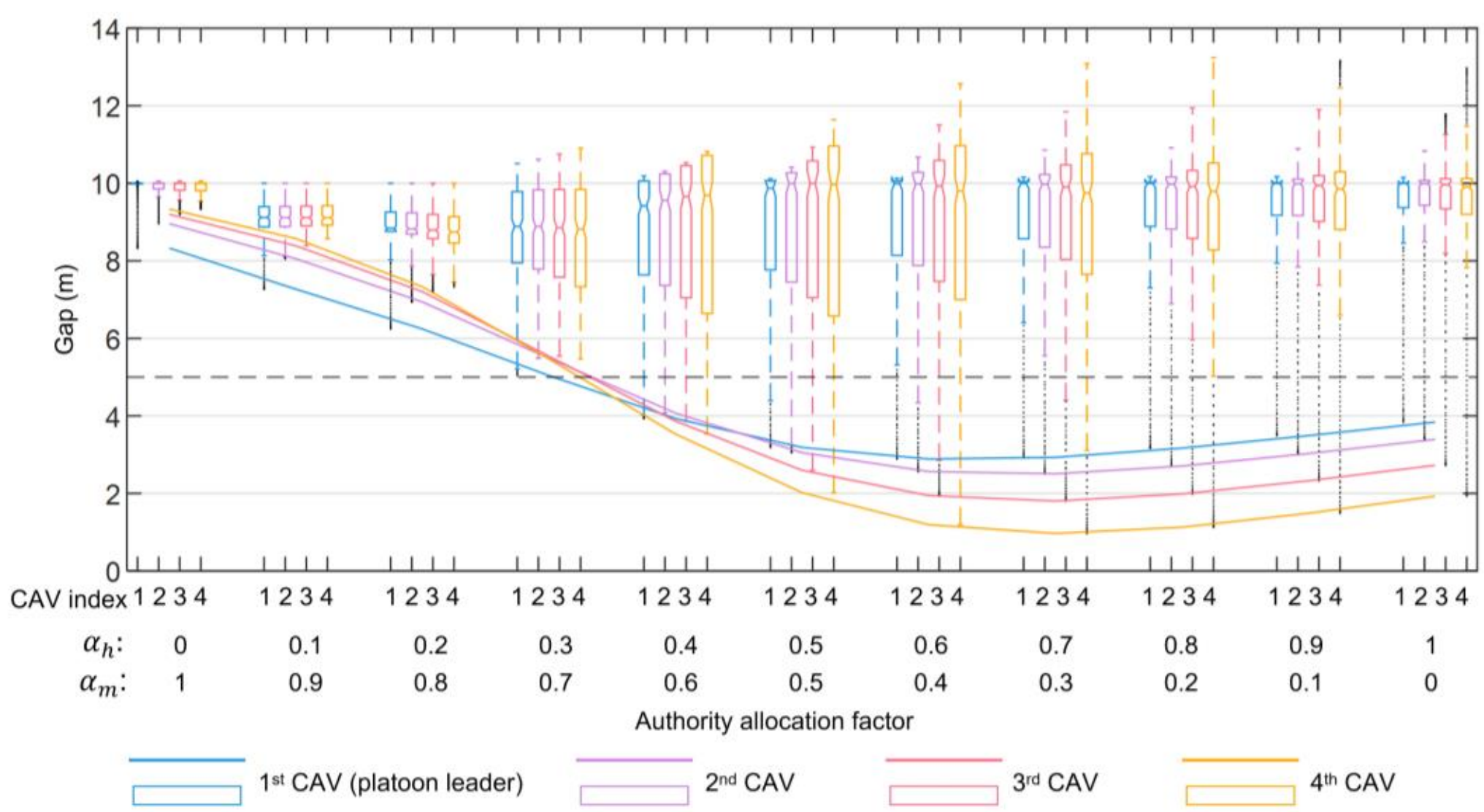

Fig. 10. Following gap distributions when leading HV conducts a hard brake.

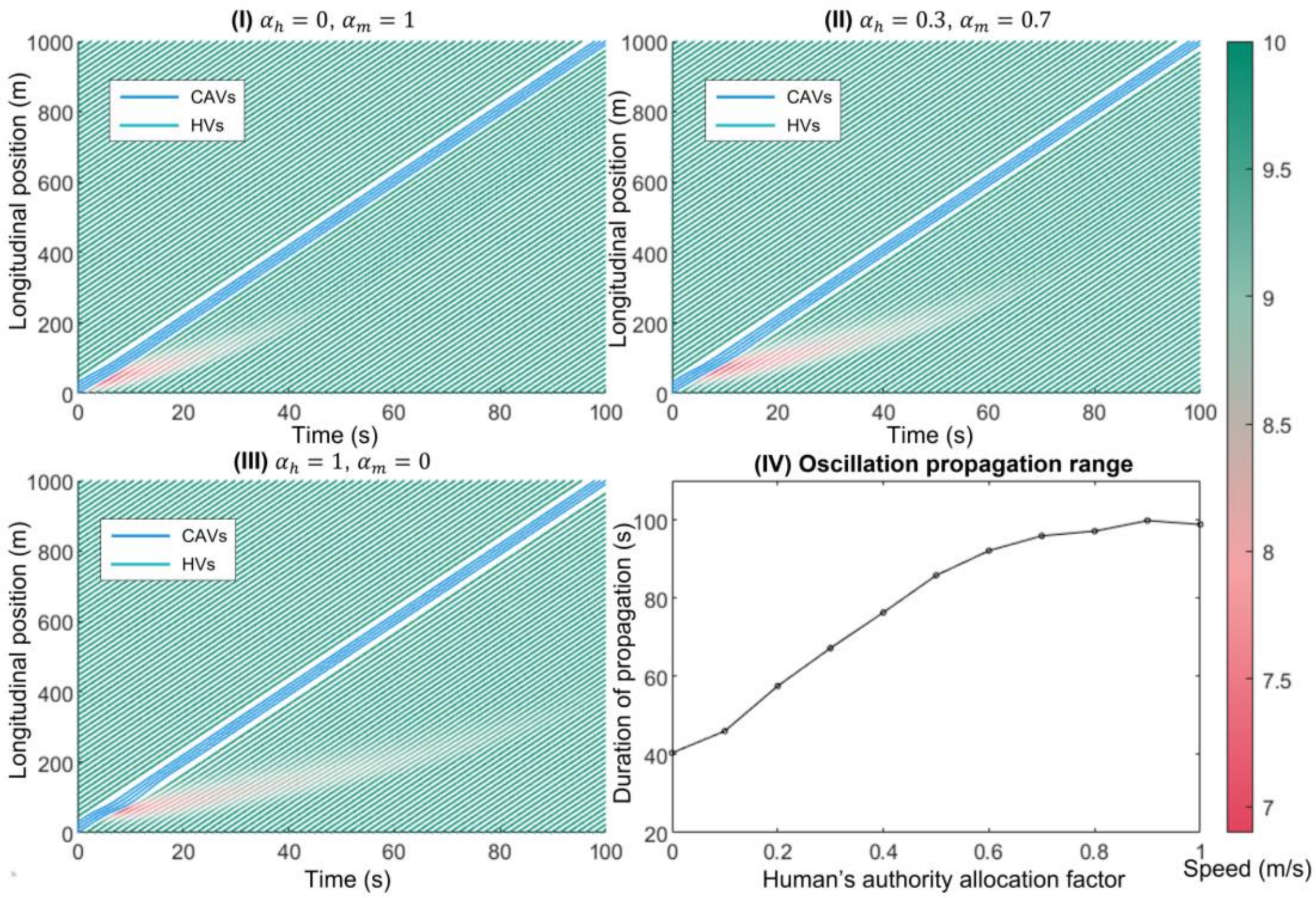

Fig. 11. Traffic influence evaluation and qualification.

*5) Actual safety*

A sensitivity analysis is conducted for actual safety evaluation regarding authority allocation factors. In the test scenario, the leading HV conducts a hard brake. The vehicles' following gap distributions and minimum following gap are illustrated in Fig. 10. Results demonstrate that the proposed controller is capable of ensuring actual safety when the human control authority is less than 0.4. In these cases, the minimum following gap is greater than the safety threshold, as illustrated in Fig. 10. However, when most of the control authorities are given to humans, collisions cannot be avoided. Hence, it is advised to restrict human control authority under 0.4 under the following gap so that human drivers could have enough reaction time. Furthermore, it could be noted that following gap oscillations increase with the increase of human's control authority, as shown in Fig. 10. When a human's control authority is less than 0.4, the minimum following gap would not increase along the platoon string. However, when human control authority increases, following gap oscillations are enhanced since the range of outliers significantly increases along the platoon string. This confirms the sting stability analysis in Fig. 8.

*6) Traffic influence*

The proposed controller is verified to reduce traffic oscillations compared to human direct takeover. The human and machine shared control scheme reduces traffic oscillation propagation, as illustrated in Fig. 11 (II), compared to entirely human control authority Fig. 11 (III). Moreover, when the machine is given greater control authority, there will be fewer traffic oscillations, as shown in Fig. 11 (I). The traffic influence range is quantified by the duration of traffic oscillation propagation, as illustrated in Fig. 11 (IV). It shows that the proposed controller is capable of reducing 60% traffic influence, by reducing the duration of traffic oscillation propagation. Moreover, traffic influence is reduced with the increase of human control authority. When the CACC platoon is fully controlled by the machine, traffic oscillation would propagate for only 40 seconds. However, when the platoon is fully controlled by humans, traffic oscillation would propagate for 100 seconds.

*7) Computational efficiency*

A sensitivity analysis is conducted in terms of prediction horizon and platoon length, as shown in Fig. 12. The evaluation is conducted on a laptop with i7-12700. Results show that the proposed controller could ensure the computing time less than 0.6s in all cases. Moreover, our evaluation is conducted in a centralized computing framework. In the actual implementation, the human-machine controller could be adopted on each CAV. In this distributed approach, the average computing time could be 0.05s, covering a prediction horizon from 2s to 20s.

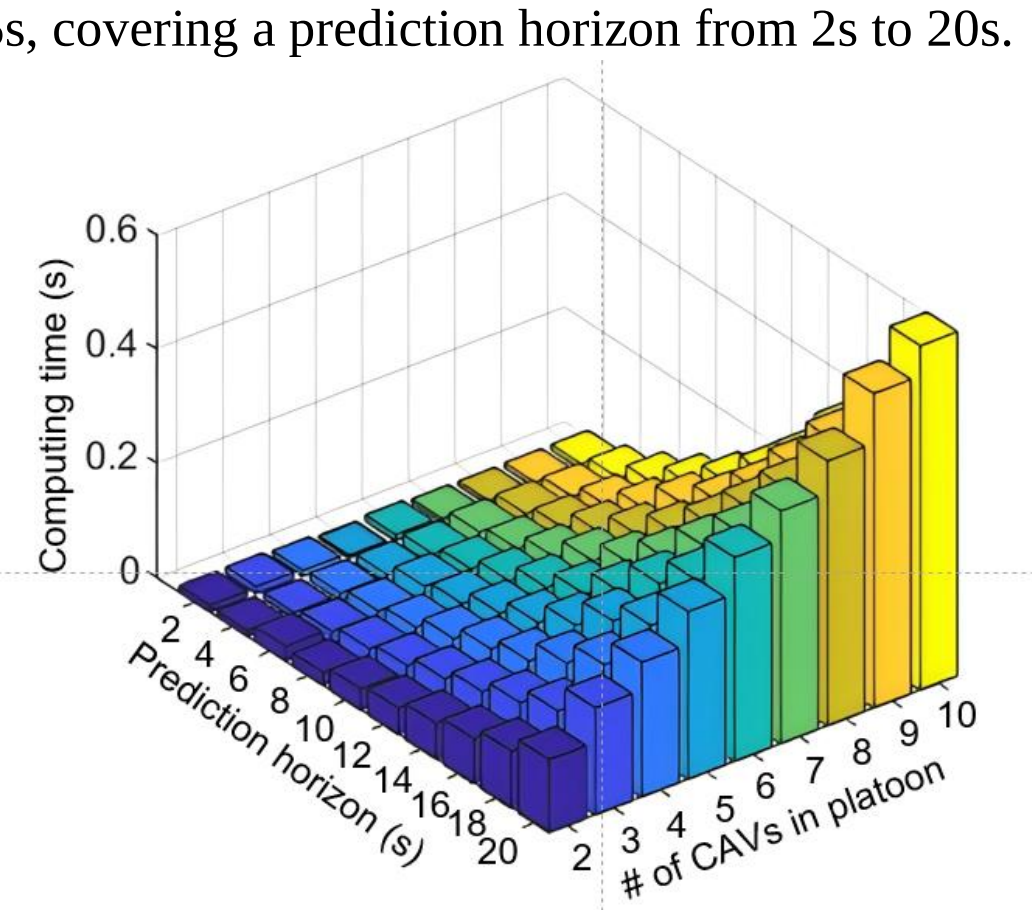

Fig. 12. Sensitivity analysis on computational efficiency.

## IV. CONCLUSIONS

This research introduces a Cooperative Adaptive Cruise Control (CACC) takeover controller based on a human-machine shared control approach, formulated as a Stackelberg competition between the machine and the human. The machine controller leads human actions by incorporating the human reaction function into the system dynamics. The proposed controller achieves the following objectives: i) enables a smooth takeover maneuver of CACC; ii) ensures string stability within a specific Operational Design Domain (ODD); iii) enhances perceived and actual safety; iv) reduces traffic oscillation propagation. Simulation results demonstrate that:

- To ensure string stability in practice, the proposed controller is suggested to be utilized in the condition that the platoon has less than 6 CAVs and human control authority is less than 40%.
- The IRL-based driving style identification method ensures 91.6% of accuracy.
- Machine interventions enhance perceived safety by at least 53.23%.
- The controller ensures actual safety when string stability is maintained.
- It reduces the influence on upstream traffic by up to 60%.
- The proposed method ensures average computing time within 0.05 seconds.

This study focuses on developing a human-machine shared control framework for CACC takeover during automation-initiated transitions. While we explicitly exclude authority allocation mechanisms for driver-initiated transitions, such cases—where drivers proactively assume control—warrant dedicated human factors analysis to optimize shared autonomy. Future work could explore adaptive arbitration strategies that harmonize driver intent with automation trajectories, informed by empirical studies of driver behavior during voluntary overrides.

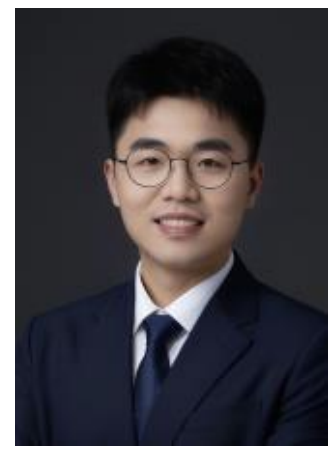

**Haoran Wang** received the bachelor's degree in transportation engineering from Tongji University, Shanghai, China, in 2017, and the Ph.D. degree from Tongji University in 2022. He is currently an Associate Professor with the College of Transportation, Tongji University. He is a researcher on vehicle engineering, majoring in intelligent vehicle control and cooperative automation.

Dr. Wang served the IEEE TRANSACTIONS ON INTELLIGENT VEHICLES, IEEE TRANSACTIONS ON INTELLIGENT TRANSPORTATION SYSTEMS, *Journal of Intelligent Transportation Systems*, and IET *Intelligent Transport Systems* as peer reviewers with a good reputation.

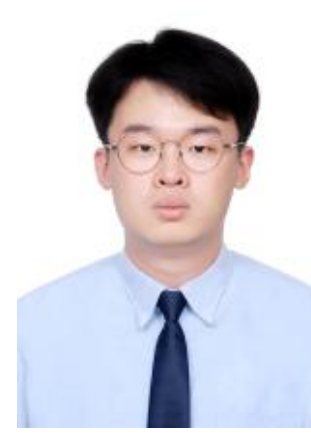

**Zhexi Lian** was born in Shanxi, China. He received the bachelor's degree in traffic engineering from College of Transportation, Tongji University, Shanghai, China, in 2023. He is currently pursuing the master's degree with Key Laboratory of Road and Traffic Engineering of the Ministry of Education, Tongji University. His main research interests include optimal control, data-driven decision-making and planning.

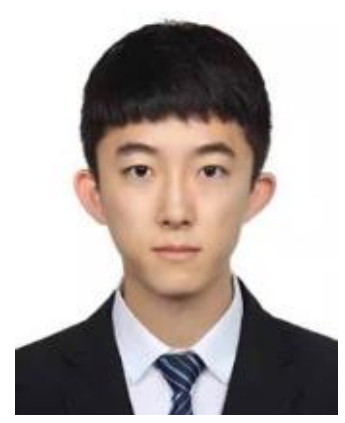

**Zhenning Li** received the B.S. and M.S. degrees in transportation science and engineering from the Harbin Institute of Technology, Harbin, China, in 2014 and 2016, respectively, and the Ph.D. degree in civil engineering from the University of Hawaii at Manoa, Honolulu, HI, USA, in 2019. He is currently an Assistant Professor with the State Key Laboratory of Internet of Things for Smart City and the Department of Computer and Information Science, University of Macau, Macau, China. His research interests include connected autonomous vehicles and Big Data application on urban transportation system.

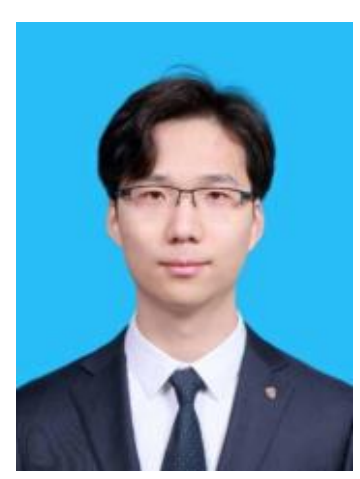

**Jiawei Wang** received the B.E. degree in Automotive Engineering from Tsinghua University, Beijing, China, in 2018, and the Ph.D. degree in Mechanical Engineering from Tsinghua University in 2023. During his Ph.D. research, he was also a visiting Ph.D. student with the Automatic Control Laboratory at École Polytechnique Fédérale de Lausanne (EPFL) in Switzerland from 2021 to 2022. He is currently a Postdoctoral Research Fellow in the Department of Civil and Environmental Engineering, University of Michigan, Ann Arbor, USA. His research interests include connected automated vehicles, intelligent vehicles, optimal control, and data-driven control.

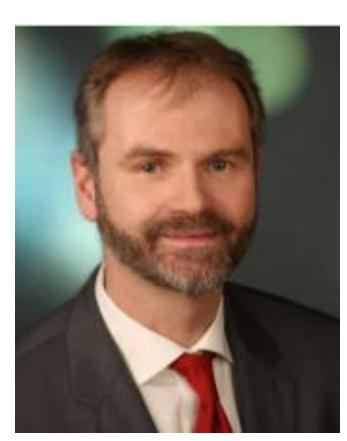

**Arno Eichberger** (Member, IEEE) received the degree in mechanical engineering and the Ph.D. degree (Hons.) in technical sciences from the Graz University of Technology, Graz, Austria, in 1995 and 1998, respectively.,From 1998 to 2007, he was a employed with Magna Steyr Fahrzeugtechnik AG&Company, Graz, where he dealt with different aspects of active and passive safety. Since 2007, he has been working with the Institute of Automotive Engineering, Graz University of Technology, dealing with driver assistance systems, vehicle dynamics, and suspensions. Since 2012, he has been an Associate Professor holding a "venia docendi" of automotive engineering.

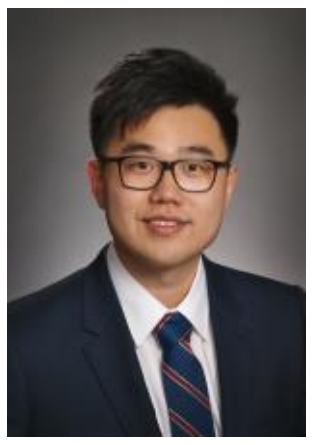

**Jia Hu** (Senior Member, IEEE) is currently working as a Zhongte Distinguished Chair of Cooperative Automation with the College of Transportation Engineering, Tongji University. Before joining Tongji University, he was a Research Associate with the Federal Highway Administration (FHWA), USA. He is an Editorial Board Member of the *Journal of Intelligent Transportation Systems* and the *International Journal of Transportation Science and Technology*. He is a member of TRB (a Division of the National Academies) Vehicle Highway Automation Commit-tee, the Freeway Operations Committee, Simulation subcommittee of Traffic Signal Systems Committee, and the Advanced Technologies Committee of the ASCE Transportation and Development Institute. He is the Chair of the Vehicle Automation and Connectivity Committee of the World Transport Convention. He is an Associate Editor of the American Society of Civil Engineers *Journal of Transportation Engineering* and IEEE OPEN JOURNAL OF INTELLIGENT TRANSPORTATION SYSTEMS.

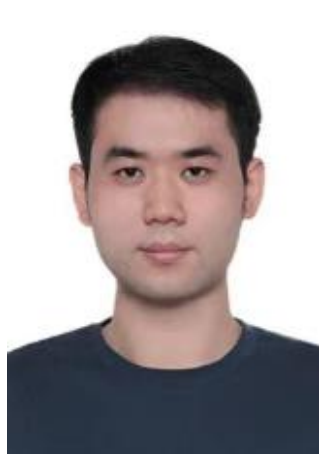

**Yongyu Chen** received a Bachelor's degree in Electrical Engineering and Automation from Shanghai University in 2013 and an M.Sc. degree in Electrical Engineering and Information Technology from the Technical University of Munich, Germany in 2017. He currently serves as a Senior Engineer at Li Auto (NASDAQ: LI), specializing in interactive decision-making for autonomous vehicles and end-to-end autonomous driving systems.

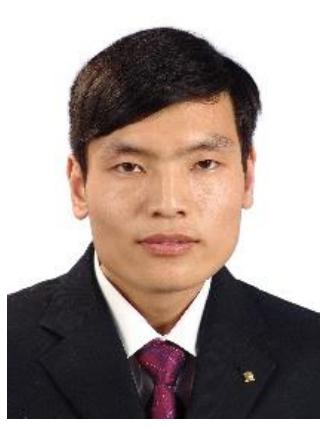

**Yongji Gao** received a Bachelor's degree in Physics from Lanzhou University in 2010 and a Master's degree in Microelectronics and Solid State Electronics from Peking University in 2014. He is currently a Director with Chery Automobile Co., Ltd. Institute of Automotive Engineering Research and Development Kaiyang Laboratory Yao Guang Laboratory Cluster Information Perception Technology Laboratory. He is an autonomous driving researcher, mainly researching perception enhancement, detection and tracking, localization and mapping, and end-to-end AI.